\documentclass{article}

 \usepackage[final]{neurips_2022}
\newcommand{\protorelated}{Prototype based models like \gls{ProtoTree}, \gls{ProtoPNet}, \gls{ProtoPShare} and \gls{ProtoPool} combine a deep feature extractor with a more interpretable model by using the similarities to learned prototypes as input to the linear model or decision tree.
While they achieve competitive performance on fine-grained image classification, \citet{kim2021hive} and \citet{hoffmann2021looks} indicate a gap between the perceived similarities of humans and prototype based models.}
\newcommand{\bldStatement}{. Best results are in bold.}
\newcommand{\firstTablespot}{}
\newcommand{\dockhorncite}{While algorithms may aim to make the models decision more interpretable (\eg applications of fuzzy models in game AI \citep{DocKru2021}), to the best of our knowledge,}
\newcommand{\rudolphcite}{In this domain, our approach could be extended to anomaly detection~\citep{Rudolph_2022_WACV} in addition to classification.}
\newcommand{\vspacehack}{\vspace{-2mm}}
\newcommand{\reinderscite}{The method could be improved via the feature selection or improvements for small datasets~\citep{chimeramix}.} %

\usepackage[utf8]{inputenc} %
\usepackage[T1]{fontenc}    %
\usepackage{url}            %
\usepackage{booktabs}       %
\usepackage{amsfonts}       %
\usepackage{nicefrac}       %
\usepackage{microtype}      %
\usepackage{xcolor}         %

\usepackage{tikz}
\usepackage[utf8]{inputenc} %
\usepackage[T1]{fontenc}    %
\usepackage{algorithm}
\usepackage{algorithmic}
\usepackage[algo2e]{algorithm2e} 
\usepackage{url}            %
\usepackage{booktabs}       %
\usepackage{amsfonts}       %
\usepackage{nicefrac}       %
\usepackage{microtype}      %
\usepackage{xcolor}         %
\usepackage{times}
\usepackage{epsfig}
\usepackage{graphicx}
\usepackage{amsmath}
\usepackage{numprint}
\usepackage{amssymb}
\usepackage{booktabs}
\usepackage{rotating}
\usepackage{makecell}
\usepackage{amssymb}%
\usepackage{pifont}%
\newcommand{\cmark}{\ding{51}}%
\newcommand{\xmark}{\ding{55}}%
\newcommand{\imgsunl}{62.2}%
\newcommand{\imgfunl}{76.7}%
\newcommand{\imgsl}{44.8}%
\newcommand{\imgfl}{72.8}%
\newcommand{\sparsemat}{\ensuremath{\gls{WeightMatrix}^{\mathrm{sparse}}}}
\newcommand{\sparsematn}[1]{\ensuremath{\gls{WeightMatrix}_#1^{\mathrm{sparse}}}}
\newcommand{\psoln}[1]{\ensuremath{(\gls{WeightMatrix}_#1^{\mathrm{sparse}},\gls{bias}_#1)}}

\newcommand{\fgvcheader}{\gls{fgvcheader}}

\newcommand{\imgnetheader}{\gls{imgnetheader}}
\newcommand{\birdsheader}{\gls{birdsheader}}
\newcommand{\inpercent}{ in percent}
\newcommand{\imgbldStatement}{ for \resnet{} on \imgnetheader{} using the pretrained dense model.}
\usepackage{svg}
\usepackage{multirow}
\usepackage{smartdiagram}
\usepackage{tikz}
\usepackage{csquotes}
\usepackage{caption}
\usepackage{arydshln}
\usepackage{subcaption}
\usetikzlibrary{shapes.geometric, arrows, fit}

\usepackage[acronym,nogroupskip]{glossaries}

\usepackage{hyperref}       %

\newglossaryentry{customLoss}
{
name={\ensuremath{\mathcal{L}_{\mathrm{div}}}},
description={Custom Loss für unterschiedliche Features}
}
\newglossaryentry{cLW}
{
name={\ensuremath{\beta}},
description={Gewichtung für customLoss}
}
\newglossaryentry{elaWeight}
{
name={\ensuremath{\lambda}},
description={Gewichtung für elasticNet}
}
\newglossaryentry{elaW}
{
name={\ensuremath{\alpha}},
long = {\ensuremath{\alpha\in[0,1]}},
description={Gewichtung zwischen l1 und l2 für elasticNet}
}
\newglossaryentry{trainDataset}
{
name={\ensuremath{\boldsymbol{D}_t}},
first = {\ensuremath{\boldsymbol{D}_t\in \mathbb{R}^{\gls{nTrainImages}\times 3\times w\times h}}},
description={LokalisierungMaps mit Missingness}
}
\newglossaryentry{nFeatures}
{
name={\ensuremath{n_{f}}},
description={Anzahl der verwendeten Features}
}
\newglossaryentry{nReducedFeatures}
{
name={\ensuremath{\gls{nFeatures}^*}},
description={Anzahl der verwendeten Features im Sparse Decision Layer}
}
\newglossaryentry{outputVector}
{
name={\ensuremath{\boldsymbol{y}}},
long = {\ensuremath{\boldsymbol{y}\in \mathbb{R}^{\gls{nClasses}}}},
description={Finaler Ausgang des Netzes}
}
\newglossaryentry{features}
{
name={\ensuremath{\boldsymbol{f}}},
description={Aus Bild berechnete Features}
}
\newglossaryentry{LocalizationMaps}
{
name={\ensuremath{\boldsymbol{L}}},
long = {\ensuremath{\boldsymbol{L}_p}\in \mathbb{R}^{\gls{nReducedFeatures}\times \frac{w}{p}\times \frac{h}{p}}},
description={LokalisierungMaps mit Missingness}
}
\newglossaryentry{featuresMapwidth}
{
name={\ensuremath{ w_M}},
description={Weite der FeatureMap}
}
\newglossaryentry{featuresMapheigth}
{
name={\ensuremath{ h_M}},
description={Weite der FeatureMap}
}
\newglossaryentry{featureMaps}
{
first ={\ensuremath{\boldsymbol{M} \in \mathbb{R}^{\gls{nFeatures} \times \gls{featuresMapwidth}\times \gls{featuresMapheigth}}}},
name={\ensuremath{m}},
plural={\ensuremath{\boldsymbol{M}}},
description={Aus Bild berechnete Features}
}
\newglossaryentry{trainFeatures}
{
first ={\ensuremath{\boldsymbol{F}^{\mathrm{train}} \in \mathbb{R}^{\gls{nTrainImages} \times \gls{nFeatures}}}},
name={\ensuremath{\boldsymbol{F}^{\mathrm{train}}}},
description={Aus Bild berechnete Features}
}
\newglossaryentry{denseNet}
{
first = {DenseNet121~\citep{huang2017densely}},
name ={DenseNet121},
description={Final layer in the neural network}
}
\newglossaryentry{resNet}
{
first = {Resnet50~\citep{he2016deep}},
name ={Resnet50},
description={Final layer in the neural network}
}
\newglossaryentry{incv}
{
first = {Inception-v3~\citep{szegedy2016rethinking}},
name ={Inception-v3},
description={Final layer in the neural network}
}
\newglossaryentry{birdsheader}
{
first = {NABirds~\citep{7298658}},
name = {NABirds},
description={Final layer in the neural network}
}
\newglossaryentry{imgnetheader}
{
first = {ImageNet-1K~\citep{imagenet15russakovsky}},
long = {ImageNet-1K~\citep{imagenet15russakovsky}},
name={ImageNet-1K},
description={Final layer in the neural network}
}
\newglossaryentry{fgvcheader}
{
first = {FGVC-Aircraft~\citep{FGVCAircraft}},
name={FGVC-Aircraft},
description={Final layer in the neural network}
}
\newglossaryentry{stanfordheader}
{
first = {Stanford Cars~\citep{StanfordCars}},
name={Stanford Cars},
description={Final layer in the neural network}
}
\newglossaryentry{cubheader}
{
first = {CUB-2011~\citep{wah2011caltech}},
long = {CUB-2011~\citep{wah2011caltech}},
name={CUB-2011},
description={Final layer in the neural network}
}
\newglossaryentry{decisionLayer}
{
name={decision layer},
description={Final layer in the neural network}
}
\newglossaryentry{fittingLossTarget}
{
name={\ensuremath{\mathcal{L}_{\mathrm{target}}}},
description={Main goal of fitting}
}
\newglossaryentry{layerName}
{
name ={\textit{SLDD-Model}},
description={The proposed benchmark}
}
\newglossaryentry{denseLayer}
{
name={{dense high-dimensional \gls{decisionLayer}}},
description={The layer that results from training}
}
\newglossaryentry{correlationMatrix}
{
name={\ensuremath{\boldsymbol{Q}}},
first = {\ensuremath{\boldsymbol{Q}\in \mathbb{R}^{\gls{nFeatures}\times\gls{nFeatures}}}},
long = {\ensuremath{q}},
description={Correlation Matrix}
}
\newglossaryentry{featureVector}
{
name={\ensuremath{f}},
long={\ensuremath{\boldsymbol{f}}},
first ={\ensuremath{\boldsymbol{f} \in \mathbb{R}^{\gls{nFeatures}}}},
description={The features of the dense alyer}
}
\newglossaryentry{RedfeatureVector}
{
name={\ensuremath{\boldsymbol{f^*}}},
first ={\ensuremath{\boldsymbol{f^*} \in \mathbb{R}^{\gls{nReducedFeatures}}}},
description={The selected feature Vector}
}
\newglossaryentry{OnlyInteractionVector}
{
name={\ensuremath{\boldsymbol{P}}},
long={\ensuremath{\boldsymbol{P} \in \mathbb{R}^{ \gls{nInteractions}}}},
description={The Interaction Vector}
}
\newglossaryentry{InteractionVector}
{
name={\ensuremath{\boldsymbol{f^*_{\phi}}}},
first ={\ensuremath{\boldsymbol{f^*_{\phi}} \in \mathbb{R}^{\gls{nReducedFeatures} + \gls{nInteractions}}}},
description={The Extended Interaction Vector}
}
\newglossaryentry{nInteractions}
{
name={\ensuremath{n_I}},
description={number of interaction term}
}
\newglossaryentry{dnn}
{
name={\ensuremath{f_\theta(x})},
description={Deep neural network}
}

\newglossaryentry{bias}
{
name={\ensuremath{\boldsymbol{b}}},
long ={\ensuremath{\boldsymbol{b} \in \mathbb{R}^{\gls{nClasses}}}},
description={The bias in the decison layer}
}
\newglossaryentry{classifyFunc}
{
name={\ensuremath{C}},
description={The classifier on the featuers}
}
\newglossaryentry{WeightMatrix}
{
name={\ensuremath{\boldsymbol{W}}},
long = {\ensuremath{\boldsymbol{W}\in \mathbb{R}^{\gls{nClasses}\times \gls{nReducedFeatures} }}},
plural = {\ensuremath{w}},
description={The Weight matrix in the decision layer}
}

\newglossaryentry{nClasses}
{
name={n_c},
description={Number of Classes}
}
\newglossaryentry{nTrainImages}
{
name={\ensuremath{n_T}},
description={Number of Train Images}
}
\newglossaryentry{nWeights}
{
name={\ensuremath{n_w}},
description={Number of Entries != 0 in \gls{WeightMatrix}}
}
\newglossaryentry{nperClass}
{
name={\ensuremath{n_{wc}}},
description={Number of Entries != 0 in \gls{WeightMatrix} per Class}
}
\newglossaryentry{interpTrans}
{
name={\ensuremath{\phi}},
description={Interpretable Transformation}
}

\newglossaryentry{targetVector}
{
name={\ensuremath{\hat{\boldsymbol{y}}}},
description={Target Vector in Training}
}
\newglossaryentry{glmsaga}
{
first = {\mbox{\textit{glm-saga~\citep{wong2021leveraging}}}},
long = {\mbox{\textit{glm-saga~\citep{wong2021leveraging}}}},
name={\mbox{\textit{glm-saga}}},
description={Target Vector in Training}
}
\newglossaryentry{cbm}
{
name={\textit{CBM}},
first ={\textit{concept bottleneck models}~(\textit{CBM})~\citep{koh2020concept}},
long = {\textit{CBM}~\citep{koh2020concept}},
description={Target Vector in Training}
}
\newglossaryentry{ProtoPNet}
{
name={\textit{ProtoPNet}},
first ={\textit{ProtoPNet}~\citep{chen2019looks}},
description={Target Vector in Training}
}
\newglossaryentry{ProtoPShare}
{
name={\textit{ProtoPShare}},
first ={\textit{ProtoPShare}~\citep{rymarczyk2021protopshare}},
description={Target Vector in Training}
}
\newglossaryentry{ProtoPool}
{
name={\textit{ProtoPool}},
first ={\textit{ProtoPool}~\citep{rymarczyk2022interpretable}},
description={Target Vector in Training}
}
\newglossaryentry{ProtoTree}
{
name={\textit{Prototree}},
first ={\textit{ProtoTree}~\citep{nauta2021neural}},
description={Target Vector in Training}
}
\newglossaryentry{ImageSample}
{
name={\ensuremath{\boldsymbol{I}}},
first ={\ensuremath{\boldsymbol{I} \in \mathbb{R}^{3\times w\times h}}},
description={The classifier on the featuers}
}
\newcommand\tFs{5 }
\newcommand{\suppt}{}%
\newcommand{\incv}{\gls{incv}}
\newcommand{\resnet}{\gls{resNet}}
\newcommand{\densenet}{\gls{denseNet}}

\newcommand{\attributeset}[1]{\ensuremath{\rho_{#1}}}
\newcommand{\classWeightsName}{\mbox{\textit{w/o Class-Specific}}}
\newcommand{\SoftmaxName}{\mbox{\textit{w/o Rescaling}}}

\newcommand{\loc}[1]{\textrm{diversity@#1}}
\newcommand{\glm}{\gls{glmsaga}}
\newcommand{\tablefinisher}[1]{ dependent on #1}
\newcommand{\undlinstmt}{ Our used \gls{cLW} is underlined.}
\newcommand{\cbmauc}{\textit{CBM-AUC}~\citep{10.1109/access.2022.3167702}}
\newcommand{\pcbm}{\textit{PCBM}~\citep{yuksekgonul2022posthoc}}

\newcommand{\eg}{\textit{e}.\textit{g}. }
\newcommand{\KeptFeatures}{\ensuremath{N_{f^*}}}
\newcommand{\initFeatures}{\ensuremath{N_{f}}}
\newcommand{\wl}[1]{\ensuremath{\boldsymbol{w}_{#1}}}
\newcommand{\st}{s.\:t.}
\newcommand{\cmpLoss}{compared to other loss functions\bldStatement }
\newcommand{\arrowDown}{\ensuremath{\boldsymbol{\downarrow}}}
\newcommand{\ourstmt}{\textbf{Ours}}
\newcommand{\emptyStrichte}{ - & - & - & - & - &}
\newcommand{\fivetext}{five}
\newcommand{\fourtext}{four}
\newcommand{\arrowUp}{\ensuremath{\boldsymbol{\uparrow}}} %
\newcommand{\elude}{\textit{Elude}~\citep{elude} }
\newcommand{\comparison}[2]{
\begin{figure*}
     \begin{subfigure}[t]{.3\textwidth}
         \centering
          \includegraphics[width=\linewidth]{plots/ComparisonSparseDenseJpg/#2/Dense#2.png}
        \caption{Conventional Dense Model}
     \end{subfigure}
     \hfill
     \begin{subfigure}[t]{.3\textwidth}
         \centering
                 \includegraphics[width=\linewidth]{plots/ComparisonSparseDenseJpg/#2/Sparse#2.png}
            \caption{Sparse with $\gls{nReducedFeatures} = \gls{nFeatures}$}
     \end{subfigure}
     \hfill
     \begin{subfigure}[t]{.3\textwidth}
         \centering
         \includegraphics[width=\linewidth]{plots/ComparisonSparseDenseJpg/#2/Finetuned#2.png}
            \caption{Finetuned SLDD-Model}
     \end{subfigure}
     \caption{Feature maps of the top 5 features by magnitude for class #1 on example images. The used weights for the respective features are also displayed. }
        \label{app:fig:vizExamples#2}
\end{figure*}
}
\tikzstyle{process} = [rectangle, minimum width=1cm, minimum height=1cm, text centered, text width=1.8cm, draw=black]
\tikzstyle{decision} = [diamond, minimum width=1cm, minimum height=1cm, text centered, draw=black]
\tikzstyle{arrow} = [thick,->,>=stealth]

\title{Take 5: Interpretable Image Classification with a Handful of Features}%

\author{%
  Thomas Norrenbrock \quad Marco Rudolph \quad Bodo Rosenhahn \\
  Institute for Information Processing (tnt)\\
  L3S - Leibniz Universität Hannover, Germany \\
  \texttt{\{norrenbr, rudolph, rosenhahn\}@tnt.uni-hannover.de}}

\begin{document}

\maketitle

\vspace{-5mm}
\begin{abstract}
\vspace{-3mm}
Deep Neural Networks use thousands of mostly incomprehensible features to identify a single class, a decision no human can follow.
We propose an interpretable sparse and low dimensional final decision layer in a deep neural network with measurable aspects of interpretability and demonstrate it on fine-grained image classification.
We argue that 
a human can only understand the decision of a machine learning model, if the features are interpretable and only very few of them are used for a single decision. For that matter, the final layer has to be sparse and - to make interpreting the features feasible - low dimensional. 
We call a model with a Sparse Low-Dimensional Decision \enquote{\gls{layerName}}.
We show that a \gls{layerName} is easier to interpret locally and globally than a \gls{denseLayer} while being able to maintain competitive accuracy. Additionally, we propose a loss function that improves a model's feature diversity and accuracy. 
Our more interpretable \gls{layerName} only uses 5 out of just 50 features per class,
while maintaining 97\,\% to 100\,\% of the accuracy on \fourtext{} common benchmark datasets compared to the baseline model with 2048 features.
\end{abstract}
\begin{figure}[bh!]
\begin{center}
   \includegraphics[width=0.5\linewidth]{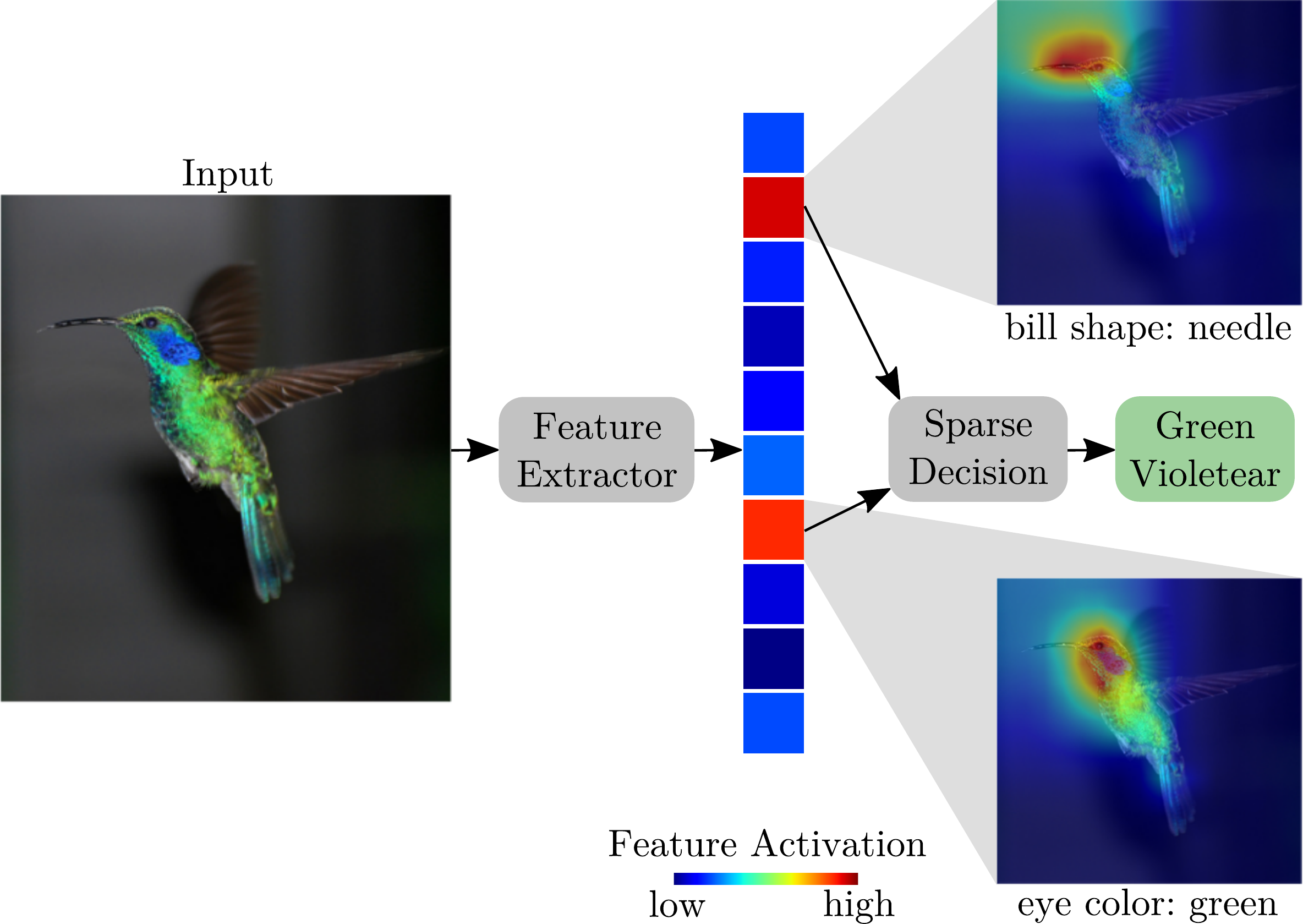}
\end{center}
   \caption{Local explanation by our \gls{layerName}: The two features used for the predicted class, emerged without additional supervision, are aligned with human interpretable attributes and localized (described in App.~\ref{app:featureViz}) adequately.}
   
\label{fig:Cover}
\end{figure}
\vspace{-10pt}
\section{Introduction}
\vspacehack{}
Understanding the decision of a deep learning model is becoming more and more important. Especially for safety-critical applications such as the medical domain or autonomous driving, it is often either legally~\citep{article} or by the practitioners required to be able to trust the decision and evaluate its reasoning~\citep{molnar2020interpretable}. 
Due to the high dimensionality of images, 
most previous work 
on interpretable models for computer vision
combines the deep features computed by a deep neural network with a method that is considered interpretable, such as a prototype based decision tree~\citep{nauta2021neural}.
While approaches for measuring the interpretability without humans exist for conventional machine learning algorithms~\citep{islam2020towards}, they are missing for methods including deep neural networks.
In this work, we propose a novel sparse and low-dimensional \gls{layerName} which offers measurable aspects of interpretability.
The key aspect is a heavily reduced number of features, out of which only very few are considered per class. Humans can only consider $7\pm 2$ aspects at once~\citep{miller1956magical} and could therefore follow a decision that uses that many features.
To be intelligible for all humans, we aim for an average of \tFs features per class. \\
 Having a reduced number of features makes it feasible to investigate every single feature and understand its meaning: We are able to align several of the learned features with human concepts post-hoc. The combination of reduced features and sparsity therefore increases both global \textit{How does the model behave?} and local interpretability \textit{Why did the model make this decision?}, demonstrated in Figure~\ref{fig:Cover}.\\
Our proposed method generates the \gls{layerName} by utilizing \glm{} to compute a sparse linear classifier for selected features, which we then finetune to the sparse structure.
We apply feature selection instead of a transformation to reduce the computational load and  preserve the original semantics of the features, which can improve interpretability~\citep{LDAForSelec}, especially if a more interpretable model like \textit{B-cos Networks}~\citep{bohle2022b} is used.
Additionally, we propose a novel loss function for more diverse features, which is especially relevant when one class depends on very few features, since using more redundant features limits the total information available for the decision.

Our main \textbf{contributions} are as follows:
\begin{itemize}

\item We present a pipeline that ensures a model with increased global and local interpretability which identifies a single class with just few, \eg~5, features of its low-dimensional representation. 
We call the resulting model \gls{layerName}.

\item Our novel feature diversity loss 
ensures diverse features. This increases the accuracy for the extremely sparse case.
    
    \item We demonstrate the competitive performance of our proposed method on four common benchmark datasets in the domain of fine-grained image classification as well as \gls{imgnetheader}, and show that several learned features for algorithmic decision-making can be directly connected to attributes humans use.%
\end{itemize}
\section{Related Work}
\vspacehack{}
\label{sec:intro}
\subsection{Fine-Grained Image Classification}
\vspacehack{}
Fine-grained image classification describes the problem of differentiating similar classes from one another. It is more challenging compared to conventional image recognition tasks~\citep{lin2015bilinear} since the differences between classes are much smaller. To tackle this difficulty, several adaptions to the common image classification approach have been applied. They usually involve learning more discriminative features by adding a term to the loss function~\citep{DevilChannels,ClassSpecificFilters,ClassUniqueCorr}, 
introducing hierarchy to the architecture~\citep{PlugInHierarchy} or using expensive expert knowledge~\citep{HierarchicalFineGrained,Flamingo}.~\citet{DevilChannels} divide the features into groups, \st{} every group is assigned to exactly one class. While training, an additional loss increases the activations of features for samples of their assigned class and reduces the overlap of feature maps in each group.~\citet{ClassSpecificFilters} tried to create class-specific filters by inducing sparsity in the features. Both~\citep{DevilChannels} and~\citep{ClassSpecificFilters} optimize for class-specific filters, which are neither suitable for the low-dimensional case when the number of classes exceeds the number of features nor interpretable, since it is unclear if the feature is already detecting the class rather than a lower level feature. The \textit{Feature Redundancy Loss}~\citep{ClassUniqueCorr} (FRL) enforces the $K$ most used features to be localized differently by reducing the normalized inner product between their feature maps. This adds a hyperparameter and does not optimize all features at once.

\subsection{Interpretable Machine Learning}
\vspacehack{}
Interpretable machine learning is a broad term and can refer to both models that are interpretable by design, and post-hoc methods that try to understand what the model has learned. Furthermore, interpretability can be classified as the interpretability of a single instance (local) or the entire model (global)~\citep{molnar2020interpretable}.\\
In this work, we present methods making models more interpretable by design but also utilize post-hoc methods to offer local and global interpretability.
Common local post-hoc methods are saliency maps like Grad-CAM~\citep{selvaraju2017grad} that aim to show what part of the input image is relevant for the prediction. While they can be helpful, they have to be cautiously interpreted, as they do not show many desired properties one would expect from an explanation like shift invariance~\citep{kindermans2019reliability} or only producing reasonable explanations, when the model is working as intended~\citep{adebayo2018sanity}. Another way of obtaining saliency maps is based on masking the input image and measuring the impact on the output~\citep{OcclusionPaper,fong2017interpretable,Missingness}.\\ %
As a global post-hoc method, \elude generates an explanation for a model by mimicking its behavior with a sparse model. %
This model uses additional attributes and main directions of the remaining feature space of the model as input. Instead of explaining a model, we directly train the more interpretable model in this work.
Another line of research tries to align learned representations with human understandable concepts from an additional labeled dataset~\citep{kim2018interpretability,bau2017network,AlphaZero}, increasing the global interpretability of the model.
\dockhorncite{}
measuring the interpretability of a deep neural network is an open task, as previous work focuses on measuring the quality of explanations of black boxes~\citep{rokade2021towards} or on conventional machine learning algorithms~\citep{islam2020towards}, where increased interpretability is measured when model complexity is reduced,
\eg via the number of operations~\citep{yang2017scalable,friedler2019assessing,ruping2006learning} or number of features~\citep{ruping2006learning}. 
The sparsity and low-dimensionality of our proposed \gls{layerName} is motivated by these findings.
Due to the limitations of post-hoc methods in explaining a deep neural network, models that are more interpretable by design are becoming more relevant.
\protorelated{}
\Gls{cbm} first predict concepts annotated in the dataset and then use a simple model to predict the target class from the concepts. 
\cbmauc{} extended \gls{cbm} by  allowing unsupervised concepts to influence the decision.
\pcbm{} created a post-hoc \gls{cbm} 
using \textit{TCAV}~\citep{kim2018interpretability} to compress high-dimensional learned features into a concept bottleneck.~\citet{margeloiu2021concept}  and \elude both suggest that training the \gls{cbm} end-to-end leads to the encoding of additional information next to the concepts, which reduces the interpretability. 
In contrast to \gls{cbm}, our proposed method does not require additional labels for training and leads to a very sparse decision process. While their features are generally more aligned with the given concepts, they also need to be analyzed thoroughly. 
\vspacehack{}
\subsubsection{\textit{Glm-saga}}
\label{sec:glmsaga}
\vspacehack{}
~\citet{wong2021leveraging} developed \glm{}, a method to efficiently fit a heavily regularized sparse layer to the computed features of a backbone feature extractor by combining the path algorithm of
\citet{friedman2010regularization} with advancements in variance reduced gradient methods by~\citet{gazagnadou2019optimal}. They showed that human understanding is more aligned with the decision process of the sparse model and sparsely shared features 
can be more easily aligned with human concepts.
Additionally, they 
reached levels of sparsity
that network wide sparsity methods do not obtain in the final
layer with competitive accuracy~\citep{StateOfSparsity}.
For precomputed and normalized features, \glm{} computes a series of $n$ sparse linear classifiers 
\begin{align}
    P= [(\sparsematn{1},\gls{bias}_1),(\sparsematn{2},\gls{bias}_2),\dots,(\sparsematn{n},\gls{bias}_n)],
\end{align}
where the sparsity of \sparsematn{i} is decreasing with $i$. This series is called \textit{regularization path}.
Each of the models minimizes the elastic net loss
\begin{align}
    \mathcal{L} = \gls{fittingLossTarget} + \gls{elaWeight} R(\gls{WeightMatrix}) &&
    R(\gls{WeightMatrix})= (1- \alpha) \frac{1}{2} \Vert \gls{WeightMatrix} \Vert_{F} + \gls{elaW} \Vert \gls{WeightMatrix} \Vert_{1,1}  \  
\end{align}
with the initial optimization goal \gls{fittingLossTarget}, in our case the cross-entropy loss, and regularization strength $\lambda$, which decreases along the path. The regularization function R(\gls{WeightMatrix}) with weighting factor \glsentrylong{elaW} is known as Elastic Net~\citep{ElasticNet}.
\Gls{glmsaga} optimizes the problem iteratively, clipping entries in \gls{WeightMatrix} with an absolute value below  a threshold
after each step, to ensure real sparsity.
For reference, the pseudocode for \glm{} is included in Appendix~\ref{app:glmsaga}.
Since their aim is understanding the neural network, they fit the sparse layer to fixed features
and do not finetune the features to the sparse layer, which requires different optimization strategies than dense networks~\citep{tessera2021keep}. 
In this work, we utilize \glm{} to create a more interpretable model with competitive accuracy by applying it on selected features with higher diversity and finetuning the features afterwards.
This leads to improved accuracy and enables a higher sparsity which, combined with the reduced number of features,increases interpretability.

\vspacehack{}
\section{Method}
\vspacehack{}
\subsection{Problem Formulation}
\vspacehack{}
We apply the proposed \gls{layerName} to the domain of fine-grained image classification.
We consider the problem of classifying an image \gls{ImageSample} of width $w$ and height $h$ into one class $c \in \{c_1,c_2, \dots, c_{\gls{nClasses}}\}$ using a trainable deep neural network $\Phi$.
This neural network extracts the feature maps \Gls{featureMaps} and
aggregates them into the feature vector \gls{featureVector}. Then it applies the trainable neural network \gls{classifyFunc} to obtain the final output \glsentrylong{outputVector} as $ \gls{outputVector} = \gls{classifyFunc}(\glsentrylong{featureVector})$.%

\subsection{\Gls{layerName}}
\vspace{-1.5mm}
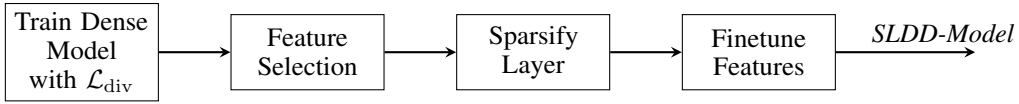
\begin{figure*}[h]
    \centering
    \begin{tikzpicture}[node distance=3cm]
\node (denseTraining) [process] {Train Dense Model with \gls{customLoss}};
\node (featureSel) [process, right of=denseTraining] {Feature Selection};
\node (fitSparse) [process, right of=featureSel] {Sparsify Layer};
\node (finetune) [process, right of=fitSparse] {Finetune Features};

\draw [arrow] (denseTraining) --   (featureSel);
\draw [arrow]  (featureSel) -- (fitSparse);
\draw [arrow] (fitSparse) --  (finetune);
\node(end) [right of=finetune] {};
\draw [arrow] (finetune) -- (end) node[midway, above, xshift=.5cm] {\gls{layerName}};
\end{tikzpicture}
    \caption{Overview of our proposed pipeline to construct a \gls{layerName}}
    \label{fig:OverviewAppraoch}
\end{figure*}
We propose a flexible, generally applicable method for generating a more locally and globally interpretable model with no need for additional labels and an adjustable tradeoff between interpretability and accuracy.
We make the decision process more interpretable by only using \gls{nReducedFeatures} features with  $\gls{nReducedFeatures} \ll \gls{nFeatures}$ and using an interpretable classifier \gls{classifyFunc}. At the core of an interpretable classifier \gls{classifyFunc} lies a linear layer $ \gls{outputVector} = \gls{WeightMatrix} \glsentrylong{featureVector} + \gls{bias}$
with the weight matrix \glsentrylong{WeightMatrix} and bias \glsentrylong{bias}. In order for it to be interpretable, \gls{WeightMatrix} has to be very sparse, meaning the number of non-zero entries  \gls{nWeights} has to be very low.~\citet{miller1956magical} showed that humans can handle $7\pm 2$ cognitive aspects at once, which constitutes an appropriate upper bound on the average number of relevant features per class $\gls{nperClass} = \frac{\gls{nWeights}}{\gls{nClasses}}$.
In our work we focus on $\gls{nperClass} \leq \tFs$.

The pipeline of our approach is presented in Figure~\ref{fig:OverviewAppraoch} and utilizes \glm{} for sparsification and feature selection.
We first train a deep neural network with our proposed feature diversity loss~\gls{customLoss} until convergence.
Then the features \Gls{trainFeatures} for all \gls{nTrainImages} images in the training set are computed,  which are the average pooled feature maps \Glspl{featureMaps}.
Afterwards, the features are selected as described in Section~\ref{sec:Selection} and \glm{}, presented in Section~\ref{sec:glmsaga}, is used to calculate the regularization path. 
Finally, the solution with the desired sparsity is selected from the regularization path and the remaining layers get finetuned with the final layer set to the sparse model, \st{} the features adapt to it.
\vspacehack{}
\subsubsection{Feature Diversity Loss}
\vspacehack{}
The goal of the proposed feature diversity loss \gls{customLoss} is that every feature captures a different independent concept. 
This is achieved by enforcing differently localized features in their feature maps.
The proposed loss is motivated by the \textit{Mutual-Channel Loss} (MCL)~\citep{DevilChannels} and the \textit{Feature Redundancy loss} (FRL)~\citep{ClassUniqueCorr}. In contrast to FRL, we use Cross-Channel-Max-Pooling (CCMP)~\citep{goodfellow2013maxout}  over all weighted feature maps to optimize all features jointly and reduce the need for the hyperparameter $K$. MCL also uses CCMP but instead of grouping the channels into class-specific filters, we apply the diversity component to all feature maps \Glspl{featureMaps}, to aim for shared interpretable features. For notation, $w_{\hat{c}l}$ describes the entry in \gls{WeightMatrix} that is assigned to the specific feature $l\in\displaystyle \{0, 1, \dots, \gls{nFeatures}-1 \}$ for the predicted class $\hat{c}= \arg\max(\gls{outputVector})\label{eq:MaxClassDiv}$ and $\gls{featureMaps}^l_{ij}=\Glspl{featureMaps}_{l,i,j}$.%
\begin{align}
\hat{s}^l_{ij} &=\frac{\exp(\gls{featureMaps}^l_{ij})}{\sum_{i'=1}^{\gls{featuresMapheigth}}\sum_{j'=1}^{\gls{featuresMapwidth}}\exp(\gls{featureMaps}^l_{i'j'})} \frac{\gls{featureVector}_l}{\max\glsentrylong{featureVector}}  \frac{|w_{\hat{c}l}|}{\Vert\boldsymbol{w}_{\hat{c}}\Vert_2} \label{eq:ScaleDiv}\\
     \gls{customLoss} &= -\sum_{i=1}^{\gls{featuresMapheigth}}\sum_{j=1}^{\gls{featuresMapwidth}}\max(\hat{s}^1_{ij},\hat{s}^2_{ij}, \dots, \hat{s}^{\gls{nFeatures}}_{ij})\label{eq:CCMPDiv}
\end{align}
Equation~\ref{eq:ScaleDiv} uses the softmax to transform the feature maps \Glspl{featureMaps} by normalizing their entries $\gls{featureMaps}^l_{ij}$ over the spatial dimensions and then %
scales the maps so that they focus on visible and important features by maintaining the relative mean of \Glspl{featureMaps} while weighting them according to the predicted class, \st{} different to MCL absent features do not have to be localized in small background patches.
Equation~\ref{eq:CCMPDiv} decreases the loss if the different weighted feature maps $\hat{S}^l$ attend to different locations.
The final training loss is then $\mathcal{L}_{\mathrm{final}} = \mathcal{L}_{CE} + \gls{cLW}\gls{customLoss}$
with the weighting factor $\gls{cLW}\in\mathbb{R}_+$.
\vspacehack{}
\subsubsection{Feature Selection}
\label{sec:Selection}
\vspacehack{}
For selecting the set of features \KeptFeatures{} from the initial features \initFeatures{} \st{} $\lvert\KeptFeatures\rvert=\gls{nReducedFeatures}$, we run an adapted version of \glm{}, introduced in Section~\ref{sec:glmsaga}, until one solution \psoln{j} of the regularization path  uses a feature not already in \KeptFeatures{}, which we then add to the set of selected features \KeptFeatures{} and restart the adapted \glm{}.
As adaptation, we extended the proximal operator of the group version of \glm{}, which operates on $\wl{l}=\gls{WeightMatrix}_{:,l}$, which are the entries in \gls{WeightMatrix} that correspond to an entire feature $l$. 
Since $\Vert\wl{l}\Vert_2$ indicates the importance of $l$, we additionally only keep entries for features that have the maximum norm or are in~\KeptFeatures{}, \st{} exactly one feature is added per iteration.
The resulting proximal operator with $\lambda_1 = \gamma\lambda\alpha$ and $\lambda_2 = \gamma\lambda(1-\alpha)$ is:
\begin{equation}
\textrm{Prox}_{\lambda_1, \lambda_2}(\wl{i}) = \begin{cases}
\frac{\wl{i}(\Vert\wl{i}\Vert_2 -\lambda_1)}{(1+\lambda_2)\Vert\wl{i}\Vert_2}
&\text{if }\Vert\wl{i}\Vert_2 > \lambda_1\underline{\land \Vert\wl{i}\Vert_2 = \max_{j'\in\initFeatures{}\setminus\KeptFeatures{}}\Vert\wl{j'}\Vert_2\lor i \in \KeptFeatures{}} \\
\boldsymbol{0} &\text{otherwise}
\end{cases}
\end{equation}
The extensions are underlined and  $\gamma$ is the learning rate of \glm{}.
\begin{table}
\begin{center}
\resizebox{\linewidth}{!}{
\centering
\begin{tabular}{c|c|c|c|c | c}
\toprule
Dataset & \glsname{cubheader} & \glsname{stanfordheader} & \glsname{fgvcheader}  & \glsname{birdsheader} & \glsname{imgnetheader}\\ 
\midrule
\#Classes $\gls{nClasses}$ & 200 & 196 & 100 & 555 & \numprint{1000}\\
Training & \numprint{5994} & \numprint{8144} & \numprint{6667}  & \numprint{23929} & \numprint{1281167}\\
Testing & \numprint{5774} & \numprint{8041} & \numprint{3333} & \numprint{24633} & \numprint{50000}\\
\bottomrule
\end{tabular}
}
\caption{Overview of the number of classes, training and testing examples for the used datasets}

\label{table:DatasetOverview}

\end{center}
\vspace{-0.38cm}
\end{table}
\begin{table}
\resizebox{\linewidth}{!}{
    \centering
    \begin{tabular}{c|c|c|c|c|c|c}
     \toprule
        Method & \thead{Additional\\Supervision required} & \thead{Features trained\\end-to-end} & \thead{\gls{nReducedFeatures} \\\arrowDown}& \thead{\gls{nperClass} \\\arrowDown} & \thead{\glsname{imgnetheader} \\\arrowUp} & \thead{\glsname{cubheader}\\\arrowUp} \\
      \midrule
        \textit{\glsentrylong{cbm} - joint} & \cmark & \cmark & 112 & 112 & - & 80.1 (76.8*)  \\
        \textit{\glsentrylong{cbm} - independent} & \cmark & \xmark & 112 & 112 & - & 76.0  \\
        \cbmauc  & \cmark & \cmark & 256 & 256 & - & \textbf{82.3} \\
        \hdashline
        \glsentrylong{glmsaga} & \xmark & \cmark & \gls{nFeatures} & $\leq \textbf{5}$ & 58.0* & 76.1 (78.0*)\\
          \gls{layerName} (\ourstmt)  with $\gls{nReducedFeatures}= \gls{nFeatures} $  & \xmark & \cmark &  \gls{nFeatures}  & \textbf{5} & \textbf{\imgfunl}* & 80.3 (\textbf{86.5*}) \\
        \gls{layerName} (\ourstmt) & \xmark & \cmark & \textbf{50} & \textbf{5} & \textbf{\imgfl}* & 78.3 (\textbf{84.0*}) \\
        \bottomrule
    \end{tabular}
    }
    \caption{Comparison with competitors on accuracy in percent. For ease of comparison, we evaluated \glsname{cubheader} on \glsname{incv} ($ \gls{nFeatures} =1024)$ and \glsname{imgnetheader} with \glsname{resNet} ($\gls{nFeatures}=2048$)  (*denotes \glsname{resNet} accuracy). The dense \glsname{resNet} achieves $76.1\,\%$ on \glsname{imgnetheader}. For \glm{}, we selected the solution  with maximum $\gls{nperClass} \leq 5$. Arrows indicate generally preferable directions. For \textit{\gls{cbm} - joint}, \glsname{resNet} results were created as described in Appendix~\ref{app:cbmjoint}}
    \label{tab:competitors}
\vspace{-0.38cm}
\end{table}
{}

\vspacehack{}
\section{Experiments}
\vspacehack{}
This section contains our experimental results. 
We validate our method using \gls{resNet}, \gls{denseNet} and \gls{incv}
on \fourtext{} common benchmark datasets in the domain of fine-grained image classification. Additionally, we show the applicability of a \gls{layerName} for large scale datasets like \glsentrylong{imgnetheader}.
An overview of \gls{cubheader}, \gls{stanfordheader}, \fgvcheader, \birdsheader{} and \imgnetheader{} is given in Table~\ref{table:DatasetOverview}.
 Additionally, \gls{cubheader} contains labels for the images such as attributes (\eg~“red wing”) as well as for the classes, which makes it easier to measure the alignment with understandable concepts.
 After the competitive accuracy and the impact of~\gls{customLoss} is shown, the interpretability of the \gls{layerName} is discussed and these attributes are used to show the alignment of the learned features.
The implementation details can be found in Appendix~\ref{app:ImpDetails}.
Finally, the tradeoff between interpretability and accuracy is visualized.

\firstTablespot{}

\subsection{Diversity Metric}
\vspacehack{}
To assess the impact of~\gls{customLoss}, we developed a measurement for the local diversity of the feature maps~\Glspl{featureMaps} that led to the decision, %
inspired by the diversity component of MCL~\citep{DevilChannels}, which entails a different way of computing the features. For that, we consider the $k$ feature maps $\Glspl{featureMaps}_k$ that are weighted the highest for the predicted class $\hat{c}$
in \gls{WeightMatrix}. 
To only compare the localization, softmax is applied to the $\Glspl{featureMaps}_k$, yielding $\boldsymbol{S}_k$.
With these distributions $\boldsymbol{S}_k$, we compute the diversity as%
\begin{equation}
    \loc{k} = \frac{\sum_{i=1}^{\gls{featuresMapheigth}}\sum_{j=1}^{\gls{featuresMapwidth}}\max(s^1_{ij},s^2_{ij}, \dots, s^k_{ij})}{k} 
\end{equation}
with $\loc{k}\in[\frac{1}{k}, 1]$ to measure
how different and pronounced
the $\Glspl{featureMaps}_k$ are localized. 
Since we focus on $\gls{nperClass} \leq 5$, we set $k=5$.
We report the mean \loc{5} for all classes that use at least \fivetext{} features.
Note that the proposed \gls{customLoss} is a weighted version of \loc{\gls{nFeatures}}.
\subsection{Results}
\label{sec:results}
\vspacehack{}
We report the accuracy on the test set for the dense model after training the pretrained model on the training data with $\gls{nperClass} = \gls{nFeatures}$, for the sparse model with $\gls{nperClass} \leq 5$, and for the result of our whole pipeline, the model with finetuned features, obtained by training the sparse model on the training data, and still $\gls{nperClass} \leq 5$.
Every shown metric is the average over \fivetext{} (\fourtext{} for \gls{imgnetheader}) randomly seeded runs. 
The standard deviations are included in the appendix.\\
Table~\ref{table:Accuracy incmp} shows 
the competitive performance of our \gls{layerName} to the dense \gls{resNet}.
It is evident that an extreme sparsity of $s = \frac{5}{2048}$
can be obtained in the final layer with just $0.1$ to $0.4$ percent points less accuracy. 
Additionally decreasing the number of features by $97.6\,\%$, resulting in just $50$ instead of the previous $2048$ features, only reduces the accuracy compared to the dense model by $1.3$ to $2.7$ percent points.
Finally, our proposed \gls{customLoss} improves the accuracy for all sparse models.
Table~\ref{table:Accuracy in backbone} shows the general applicability of our method with different backbones.
However, we observed some instability and no increased accuracy when finetuning the \gls{denseNet} with $\gls{nReducedFeatures}=2048$, showing a positive effect of sharing features. 
Table~\ref{tab:competitors} compares our approach to competitors: Without requiring additional supervision, we achieve a competitive performance compared to \glsentrylong{cbm}-based methods, while achieving a lower dimensionality and higher sparsity. Additionally, we improve the accuracy of \glm{} with heavily reduced \gls{nReducedFeatures}.
For \gls{imgnetheader}, we skipped the dense training and directly used the pretrained model.
The good scalability of our proposed method to this large dataset with a higher number of classes is displayed in Table~\ref{table:imgnetAccuracy}.
Table~\ref{table:loc5 in } shows the \loc{5}:
With \gls{customLoss} it is very close to the maximum value of $100\,\%$ in the dense case and still heavily increased in the sparse cases. This showcases that \gls{customLoss} is suitable to ensure a diverse localization and improved interpretability of the used feature maps
, which is visualized in Figures~\ref{app:fig:vizExamples0} to ~\ref{app:fig:vizExamples8}.
Finally, the total number of features that is used by the unrestricted ($\gls{nReducedFeatures} =\gls{nFeatures}$) models in Table~\ref{table:Accuracy incmp} is reduced from $912$ to $719$ for the models with~\gls{customLoss}, but still shows a high number of class-specific features.
That~\gls{customLoss} leads to more shared features supports our motivation of enforcing different features to capture different concepts.
\begin{table}
\resizebox{\linewidth}{!}{
\centering
\begin{tabular}{c|ccccc|ccccc|ccccc|ccccc}
\toprule
&\multicolumn{5}{c|}{CUB-2011}&\multicolumn{5}{c|}{FGVC-Aircraft}&\multicolumn{5}{c|}{NABirds}&\multicolumn{5}{c}{Stanford Cars}\\
 &\multicolumn{3}{c|}{$\gls{nReducedFeatures}=2048$}&\multicolumn{2}{c|}{$\gls{nReducedFeatures}=50$}& \multicolumn{3}{c|}{$\gls{nReducedFeatures}=2048$}&\multicolumn{2}{c|}{$\gls{nReducedFeatures}=50$} &
 \multicolumn{3}{c|}{$\gls{nReducedFeatures}=2048$} & \multicolumn{2}{c|}{$\gls{nReducedFeatures}=50$} &\multicolumn{3}{c|}{$\gls{nReducedFeatures}=2048$} & \multicolumn{2}{c}{$\gls{nReducedFeatures}=50$} \\
 \gls{customLoss} & Dense  & Sparse  & Finet. & Sparse  & Finet. & Dense  & Sparse  & Finet. & Sparse  & Finet. & Dense  & Sparse  & Finet. & Sparse  & Finet.  & Dense  & Sparse  & Finet. & Sparse  & Finet. \\\midrule 
\xmark & \textbf{\normalfont 86.6} & \normalfont 81.8 & \normalfont 85.3 & \normalfont 79.5 & \normalfont 83.4 & \normalfont 90.0 & \normalfont 88.4 & \normalfont 89.4 & \normalfont 87.3 & \normalfont 88.1 & \normalfont 84.2 & \normalfont 79.5 & \normalfont 83.3 & \normalfont 77.3 & \normalfont 80.7 & \normalfont 93.2 & \normalfont 90.9 & \normalfont 92.6 & \normalfont 89.3 & \normalfont 91.1 \\
\cmark & \textbf{\normalfont 86.6} & \textbf{84.0} & \textbf{86.5} & \textbf{81.7} & \textbf{84.0} & \textbf{91.4} & \textbf{90.7} & \textbf{91.1} & \textbf{89.8} & \textbf{90.1} & \textbf{84.4} & \textbf{81.0} & \textbf{84.0} & \textbf{79.8} & \textbf{81.7} & \textbf{93.6} & \textbf{92.1} & \textbf{93.3} & \textbf{91.1} & \textbf{92.0} \\
\bottomrule MCL~\cite{DevilChannels} & 86.1 & 81.9 & 85.1 & 79.4 & 82.8 & 90.1 & 88.4 & 89.0 & 87.2 & 88.1 &\emptyStrichte 93.1 & 91.0 & 92.5 & 89.0 & 90.7\\FRL~\cite{ClassUniqueCorr} & 86.4 & 81.5 & 85.3 & 78.9 & 82.6 & 90.0 & 88.5 & 89.4 & 87.5 & 88.2 &\emptyStrichte 93.3 & 90.8 & 92.6 & 89.4 & 90.9\\
\end{tabular}
}
\vspace{1mm}
\caption{Impact of the loss function on accuracy\inpercent{} for \resnet{}\bldStatement}
\label{table:Accuracy incmp}
\vspace{-5mm}
\end{table}

\begin{table}
\resizebox{\linewidth}{!}{
\centering
\begin{tabular}{c|ccccc|ccccc|ccccc|ccccc}
\toprule
&\multicolumn{5}{c|}{CUB-2011}&\multicolumn{5}{c|}{FGVC-Aircraft}&\multicolumn{5}{c|}{NABirds}&\multicolumn{5}{c}{Stanford Cars}\\
 &\multicolumn{3}{c|}{$\gls{nReducedFeatures}=\gls{nFeatures}$}&\multicolumn{2}{c|}{$\gls{nReducedFeatures}=50$}& \multicolumn{3}{c|}{$\gls{nReducedFeatures}=\gls{nFeatures}$}&\multicolumn{2}{c|}{$\gls{nReducedFeatures}=50$} &
 \multicolumn{3}{c|}{$\gls{nReducedFeatures}=\gls{nFeatures}$} & \multicolumn{2}{c|}{$\gls{nReducedFeatures}=50$} &\multicolumn{3}{c|}{$\gls{nReducedFeatures}=\gls{nFeatures}$} & \multicolumn{2}{c}{$\gls{nReducedFeatures}=50$} \\
Backbone & Dense  & Sparse  & Finet. & Sparse  & Finet. & Dense  & Sparse  & Finet. & Sparse  & Finet. & Dense  & Sparse  & Finet. & Sparse  & Finet.  & Dense  & Sparse  & Finet. & Sparse  & Finet. \\\midrule
\gls{denseNet} & 86.3 & \normalfont 76.2 & 82.9 & 75.7 & 83.1 & \textbf{91.5} & 88.2 & 89.8 & 88.1 & 90.0 & 84.1 & \normalfont 72.8 & \normalfont 64.6 & \normalfont 71.0 & 80.5 & 93.3 & \normalfont 87.3 & 91.7 & \normalfont 85.8 & 91.4 \\
\gls{incv} & \normalfont 82.3 & 78.0 & \normalfont 80.3 & \normalfont 74.0 & \normalfont 78.3 & \normalfont 88.9 & \normalfont 87.5 & \normalfont 88.1 & \normalfont 85.9 & \normalfont 87.4 & \normalfont 79.0 & 75.8 & 77.3 & 73.1 & \normalfont 76.5 & \normalfont 91.5 & 88.9 & \normalfont 90.3 & 86.3 & \normalfont 89.4 \\
\gls{resNet} & \textbf{86.6} & \textbf{84.0} & \textbf{86.5} & \textbf{81.7} & \textbf{84.0} & 91.4 & \textbf{90.7} & \textbf{91.1} & \textbf{89.8} & \textbf{90.1} & \textbf{84.4} & \textbf{81.0} & \textbf{84.0} & \textbf{79.8} & \textbf{81.7} & \textbf{93.6} & \textbf{92.1} & \textbf{93.3} & \textbf{91.1} & \textbf{92.0} \\
\bottomrule
\end{tabular}
}
\caption{Accuracy\inpercent{}\tablefinisher{backbone}\bldStatement}
\label{table:Accuracy in backbone}
\vspace{-.2cm}
\end{table}
\begin{table*}
\resizebox{\linewidth}{!}{
\centering
\begin{tabular}{ccc|cc}
\toprule
Dense ($\gls{nReducedFeatures}=2048$)  & Sparse ($\gls{nReducedFeatures}=2048$)  & Finet. ($\gls{nReducedFeatures}=2048$) & Sparse ($\gls{nReducedFeatures}=50$)  & Finet. ($\gls{nReducedFeatures}=50$) \\\midrule

76.1&\imgsunl&\imgfunl& \imgsl& \imgfl\\
\bottomrule
\end{tabular}
}
\caption{Accuracy\inpercent{}\imgbldStatement}
\label{table:imgnetAccuracy}
\vspace{-.3cm}
\end{table*}

\vspacehack{}
\subsubsection{Comparison with Other Loss Functions}
\vspacehack{}

\begin{table}
\resizebox{\linewidth}{!}{
\centering
\begin{tabular}{c|ccccc|ccccc|ccccc}
\toprule
&\multicolumn{5}{c|}{CUB-2011}&\multicolumn{5}{c|}{FGVC-Aircraft}&\multicolumn{5}{c}{Stanford Cars}\\
 &\multicolumn{3}{c|}{$\gls{nReducedFeatures}=2048$}&\multicolumn{2}{c|}{$\gls{nReducedFeatures}=50$}& \multicolumn{3}{c|}{$\gls{nReducedFeatures}=2048$}&\multicolumn{2}{c|}{$\gls{nReducedFeatures}=50$} &
 \multicolumn{3}{c}{$\gls{nReducedFeatures}=2048$} & \multicolumn{2}{c}{$\gls{nReducedFeatures}=50$}\\
 \gls{customLoss} & Dense  & Sparse  & Finet. & Sparse  & Finet. & Dense  & Sparse  & Finet. & Sparse  & Finet. & Dense  & Sparse  & Finet. & Sparse  & Finet. \\\midrule
\xmark & \normalfont 50.2 & \normalfont 46.0 & \normalfont 43.4 & \normalfont 48.0 & \normalfont 46.5 & \normalfont 46.5 & \normalfont 43.8 & \normalfont 40.9 & \normalfont 45.9 & \normalfont 44.0 & \normalfont 45.0 & \normalfont 41.7 & \normalfont 39.1 & \normalfont 43.6 & \normalfont 43.7 \\
\cmark & \textbf{98.9} & \textbf{69.9} & \textbf{71.9} & \textbf{65.2} & \textbf{72.6} & \textbf{98.8} & \textbf{85.7} & \textbf{86.6} & \textbf{69.3} & \textbf{73.9} & \textbf{99.2} & \textbf{72.4} & \textbf{74.6} & \textbf{63.7} & \textbf{74.8} \\
\bottomrule MCL~\cite{DevilChannels} & 52.5 & 51.4 & 48.9 & 56.7 & 52.3 & 50.1 & 50.6 & 48.3 & 51.7 & 50.1 & 49.0 & 49.0 & 46.2 & 51.8 & 49.5\\FRL~\cite{ClassUniqueCorr} & 51.1 & 47.1 & 44.1 & 49.0 & 46.3 & 48.2 & 44.6 & 41.2 & 44.9 & 43.1 & 46.2 & 43.0 & 40.0 & 43.0 & 41.7\\
\end{tabular}
}
\caption{Impact of the loss function on \loc{5}\inpercent{} for \resnet{}\bldStatement}
\label{table:loc5 in }
\vspace{-.3cm}
\end{table}

We compare our diversity loss \gls{customLoss} with the MCL~\citep{DevilChannels} and the FRL~\citep{ClassUniqueCorr}. 
The used hyperparameters for the loss functions are reported in Appendix~\ref{app:cbmjoint} and we focussed on three datasets to save computational resources.
Table~\ref{table:Accuracy incmp} shows that our \gls{customLoss} reaches the highest accuracy across all datasets. Notably, the accuracy reported in~\citep{DevilChannels} for the MC-Loss is achieved by a two-layer MLP plus additional techniques, whereas we only use one layer to ensure linearly separable representations for our \gls{layerName}. %
Although it is expected that applying \gls{customLoss} has a positive effect on \loc{5} due to the similar formulation, we could observe a remarkable uplift in \loc{5} (Table~\ref{table:loc5 in }) compared to MCL and FRL, which also optimize for differently localized features.

\subsection{Interpretability}%
\label{Exp:Interpretability}
\vspacehack{}
\begin{figure*}     
        \vspace{-0.45cm}
     \begin{subfigure}[t]{.47\textwidth}
         \centering
            \includegraphics[width=\textwidth]{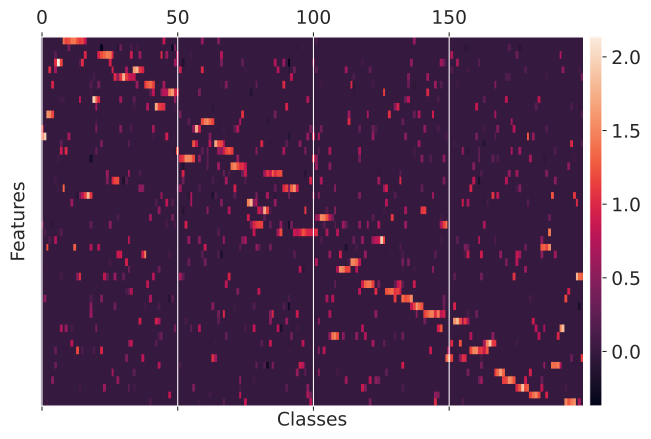}
        \caption{Exemplary \sparsemat. The alignment of the features with attributes in CUB-2011 is displayed in Figure~\ref{fig:AttMatrix}.}
        \label{fig:Matrix}
     \end{subfigure}
     \hfill
     \begin{subfigure}[t]{.47\textwidth}
         \centering
       \includegraphics[width=\textwidth]{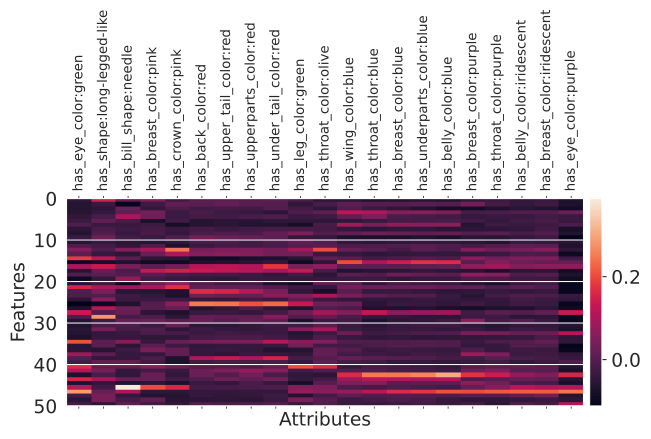}
            \caption{Relationship between chosen features and attributes ($C > 20\%)$ of CUB-2011 for the exemplary model. Higher values indicate that the feature describes the attribute.  }
        
\label{fig:AttMatrix}
     \end{subfigure}
     \caption{Visualization of the sparse matrix and the feature alignment.}
        \label{fig:matrices}
\end{figure*}
In this section, we discuss the interpretability of the proposed \gls{layerName} using example models.
The interpretability of the proposed \gls{layerName} is based upon using very few (\gls{nperClass}) features from a small pool of \gls{nReducedFeatures} to make a decision. A low \gls{nReducedFeatures} allows the analysis of the remaining features to try to align them with a human understandable concept, which is discussed in Section~\ref{sec:AlignmentFeatures}.
Since the sparse linear layer is easily interpretable%
, the complete model with sufficiently well understood features is both \textbf{locally} and \textbf{globally} interpretable.\\
For \textbf{global interpretability}, the 
final layer of the \gls{layerName} 
can be fully visualized and analyzed. Figure~\ref{fig:Matrix} shows $\gls{WeightMatrix}^{\mathrm{sparse}}$.
This allows the practitioner to verify the global behavior of the model.
For example, the attribute aligned with \enquote{has-bill-shape:needle} in the presented model in Section~\ref{sec:AlignmentFeatures} has a non-zero weight for all \fourtext{} classes that have the attribute in more than 30\,\% of examples. 
The visualization of the classes positively related to a specific attribute and features related to a class like Figures~\ref{app:fig:vizExamples0} to ~\ref{app:fig:vizExamples8} helps to trust the model.
Additionally, $\gls{WeightMatrix}^{\mathrm{sparse}}$ allows for further feature understanding, since it is possible to analyze the similarities between classes that share a feature. %
If the feature is aligned well, this leads to knowledge discovery.\\  
The \textbf{local interpretability} describes the explanation of a single decision made by the model. Decisions with sparsely connected features are inherently locally interpretable, if the features can be interpreted and localized, as shown in Figure~\ref{fig:Cover}. 
The practitioner can understand where and what was found in the image, and due to the full global interpretability also understand the behavior around the current example.

\vspacehack{}
\subsubsection{Feature Alignment}
\label{sec:AlignmentFeatures}
\vspacehack{}
In this section, we demonstrate how the features of the proposed \gls{layerName} can be aligned with interpretable concepts.%
We describe how one can use additional labels or expert knowledge to interpret the features and demonstrate that several learned sparse features are directly connected to attributes relevant to humans. Thus, our model learns such abstract concepts directly from the data.
Overall, due to the very limited number of used features \gls{nReducedFeatures}, the features can and should be thoroughly analyzed and interpreted to facilitate interpretability.
For feature localization, we follow a masking approach similar to~\citet{fong2017interpretable}, which is described in Appendix~\ref{app:featureViz}.
\paragraph{Alignment with Additional Data}
\begin{figure*}
     \begin{subfigure}[t]{.47\textwidth}
         \centering
         \includegraphics[width=\textwidth]{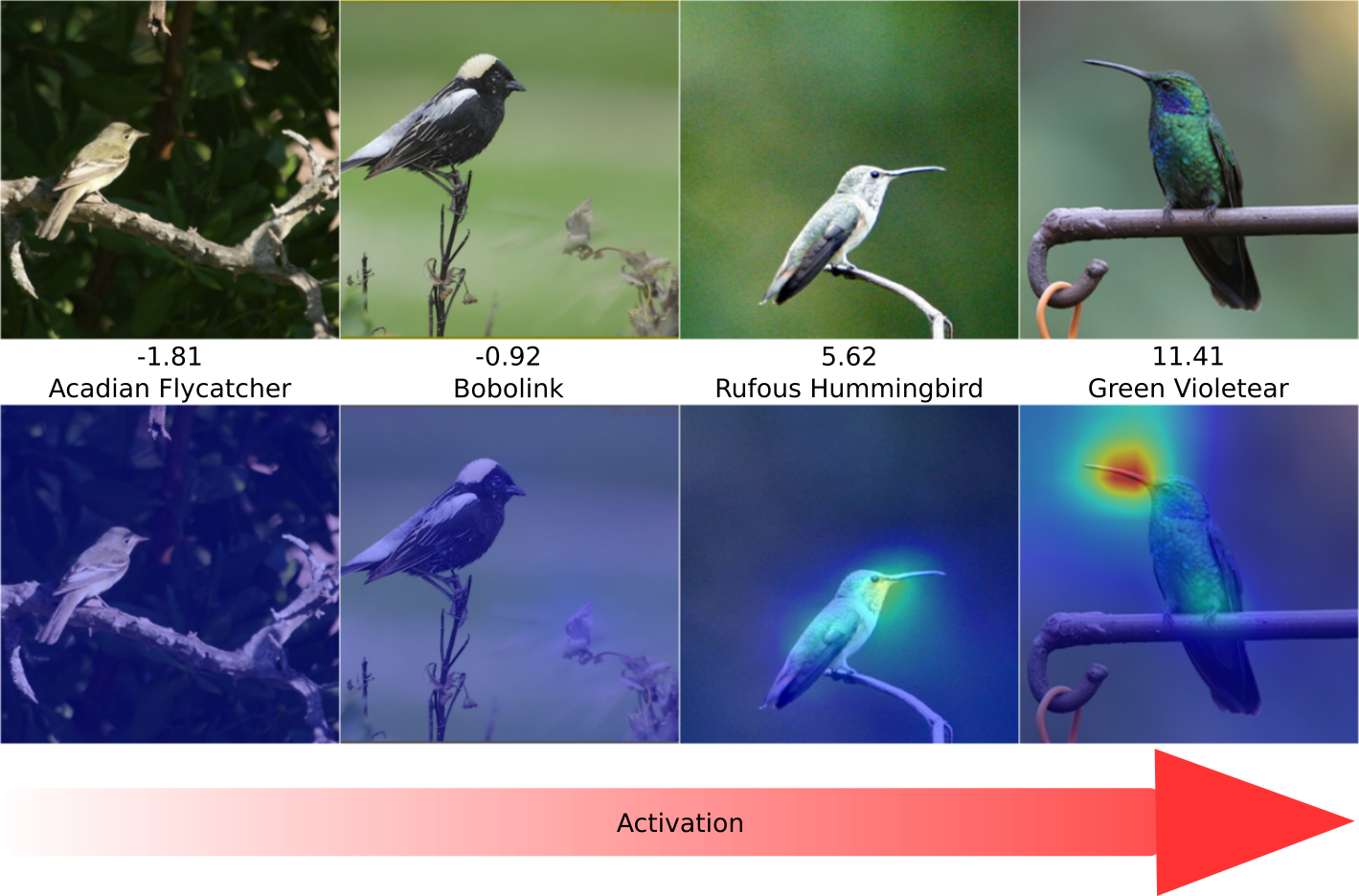}
         \caption{Feature $45$ with $C=0.39$ for the attribute \enquote{has-bill-shape:needle}: Higher activations are localized around the bill, and a needle-like bill is visible. }
         \label{fig:Schnabel}
     \end{subfigure}
     \hfill
     \begin{subfigure}[t]{.47\textwidth}
         \centering
       \includegraphics[width=\textwidth]{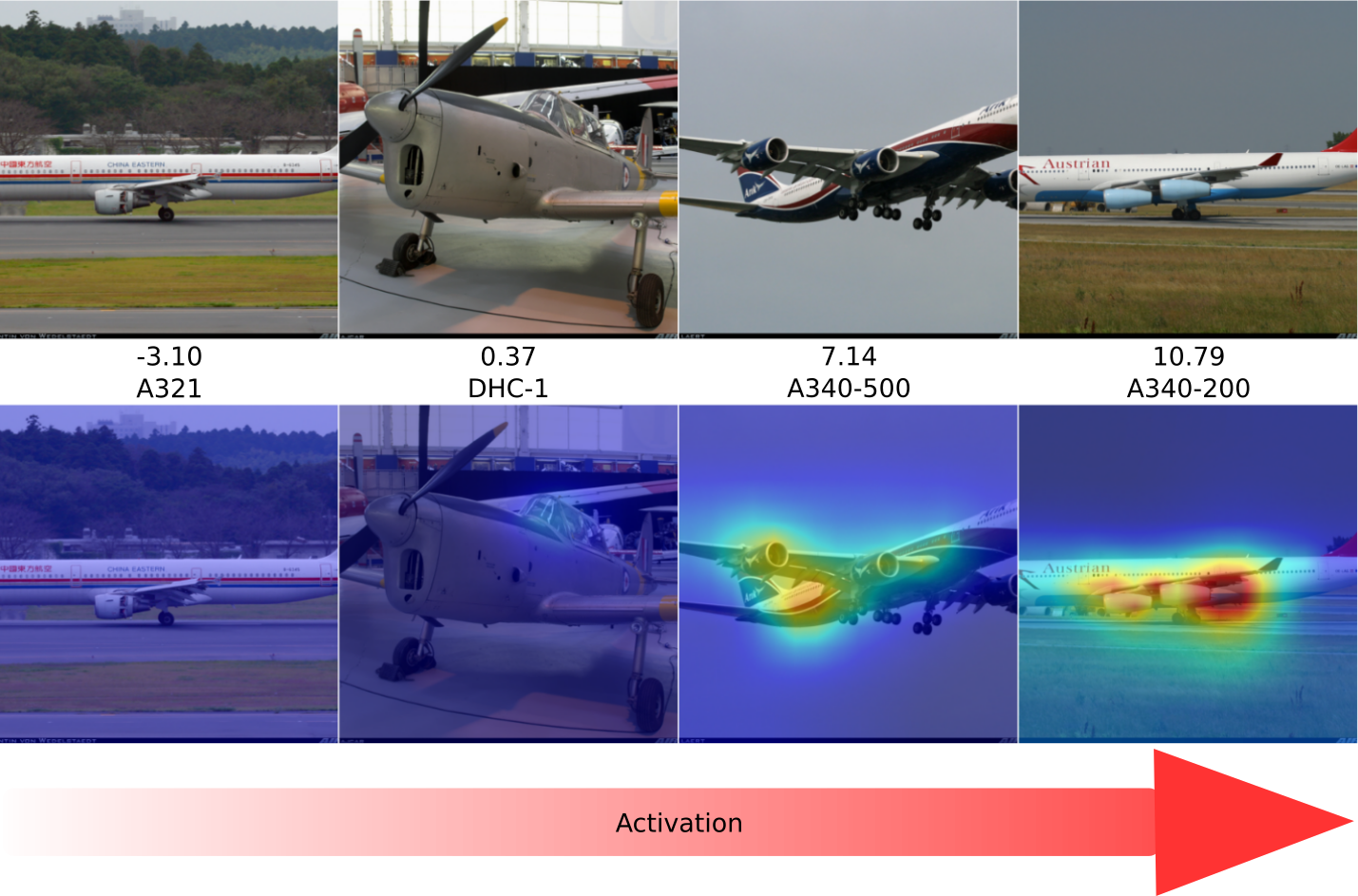}
         \caption{Manually aligned feature of a model trained on FGVC-Aircraft without additional labels: The feature is manually aligned with four engine aircraft as shown in Appendix~\ref{app:sec:alignment}.}
         \label{fig:4Strahlig}
     \end{subfigure}
        \caption{Example images and localization \gls{LocalizationMaps}, scaled to indicate feature activation, in ascending order for two models. The text between the rows describes the activation value for the image, which drops below 0 due to the normalization of \glm{}, and the class name. }
        \label{fig:three graphs}
        \vspace{-0.35cm}
\end{figure*}

We use the attributes $A$ contained in CUB-2011 to align the learned features with these labels after the finetuning. 
For each attribute $a \in A$ and feature $j$ we compute a score $C_{aj}$ that corresponds to a relative increase of the feature when the attribute is present:
\begin{align}
\delta_{aj} = \frac{1}{\lvert\attributeset{a+}\rvert}\sum_{i\in\attributeset{a+}}\gls{trainFeatures}_{i,j}- \frac{1}{\lvert\attributeset{a-}\rvert}\sum_{i\in\attributeset{a-}}\gls{trainFeatures}_{i,j} &&
    C_{aj} = \frac{\delta_{aj}}{\max(\gls{trainFeatures}_{:,j}) - \min(\gls{trainFeatures}_{:,j})} 
\end{align}
The set of indices whose images contain the attribute is denoted by~\attributeset{a+}.
We considered an attribute to be present if the human annotated it with \enquote{probably} or \enquote{definitely}.  Annotations with \enquote{guessing} were neither included in the positive (\attributeset{a+}) nor negative (\attributeset{a-}) examples.
For one exemplary model a part of the matrix of $C$ values is displayed in Figure~\ref{fig:AttMatrix}. 
It is clear, that some features correspond to colors, some to specific shapes like “bill-shape:needle” and other features do not correlate with specific attributes. 
Figure~\ref{fig:Schnabel} visually validates the connection that was implied in Figure~\ref{fig:AttMatrix}.
\paragraph{Manual Alignment}
To ensure that the entirety of a feature is understood, or in absence of additional data, the features have to be manually aligned for increased interpretability. This is enabled by the low number of features and sparsity.
Some useful aspects for understanding a feature are the localization, extreme examples, feature visualization~\citep{olah2017feature} or which classes use that feature. One such alignment for a model trained on FGVC-Aircraft is displayed in Appendix~\ref{app:sec:alignment} and Figure~\ref{fig:4Strahlig}.
Aligning learned features with human understandable concepts is still challenging as a single feature can refer to multiple aspects and human understandable concepts do not need to be axis aligned~\citep{szegedy2013intriguing}. However, the low dimensionality of the remaining features allows for a sophisticated analysis of every feature in practice, which could even discover spurious correlations as features as done by the \glm{} publication.
\subsection{Interpretability Tradeoff}
\vspacehack{}

\begin{figure}
\centering
\hspace{0.05\linewidth}
     \begin{subfigure}[t]{.4\linewidth}
         \centering
         \includegraphics[width=\textwidth]{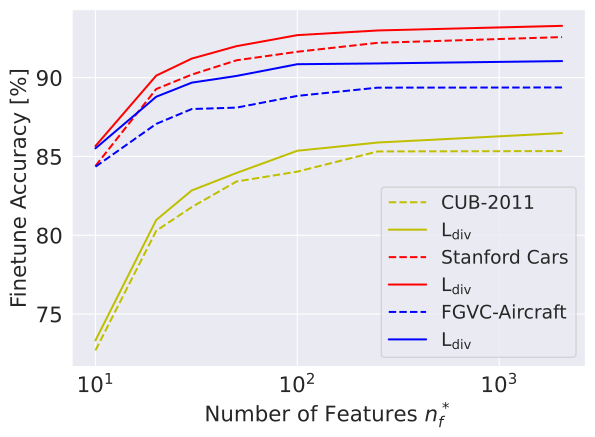}
          
         \caption{Impact of changing \gls{nReducedFeatures} with $\gls{nperClass}=5$ }
         \label{fig:NFeaturesImpact}
     \end{subfigure}
     \hfill
     \begin{subfigure}[t]{.4\linewidth}
         \centering
         
       \includegraphics[width=\textwidth]{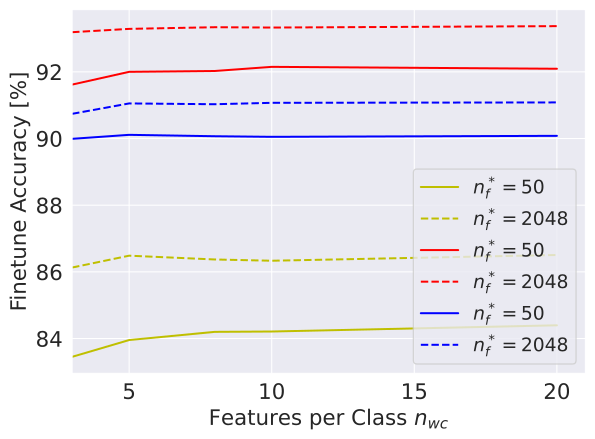}
       
         \caption{Impact of changing \gls{nperClass} with $\gls{nReducedFeatures}=50$ or $2048$ }
         \label{fig:Impactnwc}
         
     \end{subfigure}
     \hspace{0.05\linewidth}
        \caption{Relationship between Finetune Accuracy and aspects related to interpretability for \gls{resNet}. }
        \label{fig:ContinousInt}
    \vspace{-.2cm}
\end{figure}

In this section, we analyze the impact of changing \gls{nReducedFeatures} and \gls{nperClass} on the finetuning accuracy of the model trained with \gls{customLoss}, shown in Figure~\ref{fig:ContinousInt}. 
Figure~\ref{fig:NFeaturesImpact} visualizes the impact of \gls{nReducedFeatures}: 
With decreasing \gls{nReducedFeatures} the accuracy drops slowly until a dataset-specific threshold is reached, at which a steep decline starts.
Additionally, the proposed~\gls{customLoss} works regardless of~\gls{nReducedFeatures}. 
Figure~\ref{fig:Impactnwc} shows the finetuning accuracy in relation to \gls{nperClass}: 
The accuracy is rather insensitive to \gls{nperClass},
only decreasing when $\gls{nperClass}<5$, which is the case for both dimensions and showcases that \fivetext{} features suffice for a competitive model even if the features are shared among classes.
Figure~\ref{fig:ContinousInt} demonstrates the tradeoff that our \gls{layerName} offers: Both \gls{nperClass} and \gls{nReducedFeatures} can be drastically reduced with either a negligible or small impact on accuracy to adapt to the amount of interpretability needed.

\vspacehack{}
\section{Limitations and Future Work}
\label{sec:limits}
\vspacehack{}
As shown in Figure~\ref{fig:ContinousInt}, the \gls{layerName} cannot get arbitrarily low-dimensional or sparse with competitive accuracy via our proposed method. 
The optimal sparsity and dimensionality for a given problem are hard to predict and might require some experiments to determine the minimum values for competitive accuracy.
Aligning all used features with human concepts is still difficult, albeit more feasible than without a \gls{layerName}.
Future work could use a more interpretable feature extractor like \textit{B-cos Networks}~\citep{bohle2022b} to alleviate that problem.
\reinderscite{}
It seems promising to apply a \gls{layerName} to other safety-critical domains, such as medical, where an expert can be utilized to align the features and follow the decision, as it can help bring the required interpretability and trustworthiness to the domain.
Embodied autonomous agents can also benefit from it, as the entire decision process can be thoroughly analyzed.
\rudolphcite{}
While more interpretable models could be used to more deliberately bring harm, they can disclose existing problems with machine learning models and open up the opportunity to build fair and trustworthy models.
Finally, sparsity and dimensionality could be part of metrics used to quantify the trustworthiness of a model.
\section{Conclusion}
\vspacehack{}
In this work, we proposed the more interpretable sparse low-dimensional decision model (\gls{layerName})  to allow a human to follow and understand the decision of a Deep Neural Network for image classification.
Our proposed pipeline constructs a \gls{layerName} with drastically increased global and local interpretability while still showing competitive accuracy.
As demonstrated, a practitioner can manually configure the pipeline to set the tradeoff between accuracy and interpretability.
Our novel loss increases the feature diversity and we showed that identifying a class with varied features can improve the accuracy. 
Finally, our \gls{layerName} offers measurable aspects of interpretability, which allows future work to not just compare itself on accuracy but also on interpretability.

\vspace{-0.1em}
\small{\paragraph{Acknowledgements.}
This work was supported by the Federal Ministry of Education and
Research (BMBF), Germany under the project LeibnizKILabor (grant no.
01DD20003), the Center for Digital Innovations (ZDIN) and the Deutsche Forschungsgemeinschaft  (DFG) under  Germany’s  Excellence  Strategy  within  the  Cluster of Excellence PhoenixD (EXC 2122).

\bibliography{egbib}

\begin{thebibliography}{10}\itemsep=-1pt

\bibitem[Adebayo et~al.(2018)]{adebayo2018sanity}
Julius Adebayo, Justin Gilmer, Michael Muelly, Ian Goodfellow, Moritz Hardt,
  and Been Kim.
\newblock Sanity checks for saliency maps.
\newblock {\em Advances in neural information processing systems}, 31, 2018.

\bibitem[Bau et~al.(2017)]{bau2017network}
David Bau, Bolei Zhou, Aditya Khosla, Aude Oliva, and Antonio Torralba.
\newblock Network dissection: Quantifying interpretability of deep visual
  representations.
\newblock In {\em Proceedings of the IEEE conference on computer vision and
  pattern recognition}, pages 6541--6549, 2017.

\bibitem[Bibal et~al.(2021)]{article}
Adrien Bibal, Michael Lognoul, Alexandre Streel, and Benoît Frénay.
\newblock Legal requirements on explainability in machine learning.
\newblock {\em Artificial Intelligence and Law}, 29, 06 2021.

\bibitem[B{\"o}hle et~al.(2022)]{bohle2022b}
Moritz B{\"o}hle, Mario Fritz, and Bernt Schiele.
\newblock B-cos networks: Alignment is all we need for interpretability.
\newblock In {\em Proceedings of the IEEE/CVF Conference on Computer Vision and
  Pattern Recognition}, pages 10329--10338, 2022.

\bibitem[Chang et~al.(2020)]{DevilChannels}
Dongliang Chang, Yifeng Ding, Jiyang Xie, Ayan~Kumar Bhunia, Xiaoxu Li, Zhanyu
  Ma, Ming Wu, Jun Guo, and Yi-Zhe Song.
\newblock {The Devil is in the Channels: Mutual-Channel Loss for Fine-Grained
  Image Classification}.
\newblock {\em IEEE Transactions on Image Processing}, 29:4683--4695, 2020.

\bibitem[Chang et~al.(2021)]{Flamingo}
Dongliang Chang, Kaiyue Pang, Yixiao Zheng, Zhanyu Ma, Yi-Zhe Song, and Jun
  Guo.
\newblock {Your “Flamingo” is My “Bird”: Fine-Grained, or Not}.
\newblock {\em 2021 IEEE/CVF Conference on Computer Vision and Pattern
  Recognition (CVPR)}, 00:11471--11480, 2021.

\bibitem[Chen et~al.(2019)]{chen2019looks}
Chaofan Chen, Oscar Li, Daniel Tao, Alina Barnett, Cynthia Rudin, and
  Jonathan~K Su.
\newblock This looks like that: deep learning for interpretable image
  recognition.
\newblock {\em Advances in neural information processing systems}, 32, 2019.

\bibitem[Chen et~al.(2018)]{HierarchicalFineGrained}
Tianshui Chen, Wenxi Wu, Yuefang Gao, Le Dong, Xiaonan Luo, and Liang Lin.
\newblock Fine-grained representation learning and recognition by exploiting
  hierarchical semantic embedding.
\newblock In {\em Proceedings of the 26th ACM international conference on
  Multimedia}, pages 2023--2031, 2018.

\bibitem[Chou et~al.(2022)]{PlugInHierarchy}
Po-Yung Chou, Cheng-Hung Lin, and Wen-Chung Kao.
\newblock {A Novel Plug-in Module for Fine-Grained Visual Classification}.
\newblock {\em arXiv}, 2022.

\bibitem[Dockhorn and Kruse(2021)]{DocKru2021}
Alexander Dockhorn and Rudolf Kruse.
\newblock Fuzzy modeling in game ai.
\newblock {\em Journal of Pure and Applied Mathematics}, 12(1):54--68, 2021.

\bibitem[Engstrom et~al.(2019)]{robustness}
Logan Engstrom, Andrew Ilyas, Hadi Salman, Shibani Santurkar, and Dimitris
  Tsipras.
\newblock Robustness (python library), 2019.

\bibitem[Fong and Vedaldi(2017)]{fong2017interpretable}
Ruth~C Fong and Andrea Vedaldi.
\newblock Interpretable explanations of black boxes by meaningful perturbation.
\newblock In {\em Proceedings of the IEEE international conference on computer
  vision}, pages 3429--3437, 2017.

\bibitem[Friedler et~al.(2019)]{friedler2019assessing}
Sorelle~A Friedler, Chitradeep~Dutta Roy, Carlos Scheidegger, and Dylan Slack.
\newblock Assessing the local interpretability of machine learning models.
\newblock 2019.

\bibitem[Friedman et~al.(2010)]{friedman2010regularization}
Jerome Friedman, Trevor Hastie, and Rob Tibshirani.
\newblock Regularization paths for generalized linear models via coordinate
  descent.
\newblock {\em Journal of statistical software}, 33(1):1, 2010.

\bibitem[Gale et~al.(2019)]{StateOfSparsity}
Trevor Gale, Erich Elsen, and Sara Hooker.
\newblock The state of sparsity in deep neural networks.
\newblock {\em arXiv preprint arXiv:1902.09574}, 2019.

\bibitem[Gazagnadou et~al.(2019)]{gazagnadou2019optimal}
Nidham Gazagnadou, Robert Gower, and Joseph Salmon.
\newblock Optimal mini-batch and step sizes for saga.
\newblock In {\em International conference on machine learning}, pages
  2142--2150. PMLR, 2019.

\bibitem[Goodfellow et~al.(2013)]{goodfellow2013maxout}
Ian Goodfellow, David Warde-Farley, Mehdi Mirza, Aaron Courville, and Yoshua
  Bengio.
\newblock Maxout networks.
\newblock In {\em International conference on machine learning}, pages
  1319--1327. PMLR, 2013.

\bibitem[He et~al.(2016)]{he2016deep}
Kaiming He, Xiangyu Zhang, Shaoqing Ren, and Jian Sun.
\newblock Deep residual learning for image recognition.
\newblock In {\em Proceedings of the IEEE conference on computer vision and
  pattern recognition}, pages 770--778, 2016.

\bibitem[Hoffmann et~al.(2021)]{hoffmann2021looks}
Adrian Hoffmann, Claudio Fanconi, Rahul Rade, and Jonas Kohler.
\newblock This looks like that... does it? shortcomings of latent space
  prototype interpretability in deep networks.
\newblock {\em arXiv preprint arXiv:2105.02968}, 2021.

\bibitem[Huang et~al.(2017)]{huang2017densely}
Gao Huang, Zhuang Liu, Laurens Van Der~Maaten, and Kilian~Q Weinberger.
\newblock Densely connected convolutional networks.
\newblock In {\em Proceedings of the IEEE conference on computer vision and
  pattern recognition}, pages 4700--4708, 2017.

\bibitem[Islam et~al.(2020)]{islam2020towards}
Sheikh~Rabiul Islam, William Eberle, and Sheikh~K Ghafoor.
\newblock Towards quantification of explainability in explainable artificial
  intelligence methods.
\newblock In {\em The thirty-third international flairs conference}, 2020.

\bibitem[Jain et~al.(2022)]{Missingness}
Saachi Jain, Hadi Salman, Eric Wong, Pengchuan Zhang, Vibhav Vineet, Sai
  Vemprala, and Aleksander Madry.
\newblock Missingness bias in model debugging.
\newblock In {\em International Conference on Learning Representations}, 2022.

\bibitem[Kim et~al.(2018)]{kim2018interpretability}
Been Kim, Martin Wattenberg, Justin Gilmer, Carrie Cai, James Wexler, Fernanda
  Viegas, et~al.
\newblock Interpretability beyond feature attribution: Quantitative testing
  with concept activation vectors (tcav).
\newblock In {\em International conference on machine learning}, pages
  2668--2677. PMLR, 2018.

\bibitem[Kim et~al.(2021)]{kim2021hive}
Sunnie~SY Kim, Nicole Meister, Vikram~V Ramaswamy, Ruth Fong, and Olga
  Russakovsky.
\newblock Hive: evaluating the human interpretability of visual explanations.
\newblock {\em arXiv preprint arXiv:2112.03184}, 2021.

\bibitem[Kindermans et~al.(2019)]{kindermans2019reliability}
Pieter-Jan Kindermans, Sara Hooker, Julius Adebayo, Maximilian Alber, Kristof~T
  Sch{\"u}tt, Sven D{\"a}hne, Dumitru Erhan, and Been Kim.
\newblock The (un) reliability of saliency methods.
\newblock In {\em Explainable AI: Interpreting, Explaining and Visualizing Deep
  Learning}, pages 267--280. Springer, 2019.

\bibitem[Koh et~al.(2020)]{koh2020concept}
Pang~Wei Koh, Thao Nguyen, Yew~Siang Tang, Stephen Mussmann, Emma Pierson, Been
  Kim, and Percy Liang.
\newblock Concept bottleneck models.
\newblock In {\em International Conference on Machine Learning}, pages
  5338--5348. PMLR, 2020.

\bibitem[Krause et~al.(2013)]{StanfordCars}
Jonathan Krause, Michael Stark, Jia Deng, and Li Fei-Fei.
\newblock 3d object representations for fine-grained categorization.
\newblock In {\em 4th International IEEE Workshop on 3D Representation and
  Recognition (3dRR-13)}, Sydney, Australia, 2013.

\bibitem[Liang et~al.(2020)]{ClassSpecificFilters}
Haoyu Liang, Zhihao Ouyang, Yuyuan Zeng, Hang Su, Zihao He, Shu-Tao Xia, Jun
  Zhu, and Bo Zhang.
\newblock Training interpretable convolutional neural networks by
  differentiating class-specific filters.
\newblock In {\em European Conference on Computer Vision}, pages 622--638.
  Springer, 2020.

\bibitem[Lin et~al.(2015)]{lin2015bilinear}
Tsung-Yu Lin, Aruni RoyChowdhury, and Subhransu Maji.
\newblock Bilinear cnn models for fine-grained visual recognition.
\newblock In {\em Proceedings of the IEEE international conference on computer
  vision}, pages 1449--1457, 2015.

\bibitem[Maji et~al.(2013)]{FGVCAircraft}
S. Maji, J. Kannala, E. Rahtu, M. Blaschko, and A. Vedaldi.
\newblock Fine-grained visual classification of aircraft.
\newblock Technical report, 2013.

\bibitem[Margeloiu et~al.(2021)]{margeloiu2021concept}
Andrei Margeloiu, Matthew Ashman, Umang Bhatt, Yanzhi Chen, Mateja Jamnik, and
  Adrian Weller.
\newblock Do concept bottleneck models learn as intended?
\newblock {\em arXiv preprint arXiv:2105.04289}, 2021.

\bibitem[McGrath et~al.(2021)]{AlphaZero}
Thomas McGrath, Andrei Kapishnikov, Nenad Tomašev, Adam Pearce, Demis
  Hassabis, Been Kim, Ulrich Paquet, and Vladimir Kramnik.
\newblock {Acquisition of Chess Knowledge in AlphaZero}.
\newblock {\em arXiv}, 2021.

\bibitem[Miller(1956)]{miller1956magical}
George~A Miller.
\newblock The magical number seven, plus or minus two: Some limits on our
  capacity for processing information.
\newblock {\em Psychological review}, 63(2):81, 1956.

\bibitem[Molnar(2020)]{molnar2020interpretable}
Christoph Molnar.
\newblock {\em Interpretable machine learning}.
\newblock Lulu. com, 2020.

\bibitem[Nauta et~al.(2021)]{nauta2021neural}
Meike Nauta, Ron van Bree, and Christin Seifert.
\newblock Neural prototype trees for interpretable fine-grained image
  recognition.
\newblock In {\em Proceedings of the IEEE/CVF Conference on Computer Vision and
  Pattern Recognition}, pages 14933--14943, 2021.

\bibitem[Olah et~al.(2017)]{olah2017feature}
Chris Olah, Alexander Mordvintsev, and Ludwig Schubert.
\newblock Feature visualization.
\newblock {\em Distill}, 2(11):e7, 2017.

\bibitem[Paszke et~al.(2019)]{Pytorch}
Adam Paszke, Sam Gross, Francisco Massa, Adam Lerer, James Bradbury, Gregory
  Chanan, Trevor Killeen, Zeming Lin, Natalia Gimelshein, Luca Antiga, Alban
  Desmaison, Andreas Kopf, Edward Yang, Zachary DeVito, Martin Raison, Alykhan
  Tejani, Sasank Chilamkurthy, Benoit Steiner, Lu Fang, Junjie Bai, and Soumith
  Chintala.
\newblock Pytorch: An imperative style, high-performance deep learning library.
\newblock In H. Wallach, H. Larochelle, A. Beygelzimer, F. d\textquotesingle
  Alch\'{e}-Buc, E. Fox, and R. Garnett, editors, {\em Advances in Neural
  Information Processing Systems 32}, pages 8024--8035. Curran Associates,
  Inc., 2019.

\bibitem[Ramaswamy et~al.(2022)]{elude}
Vikram~V Ramaswamy, Sunnie S~Y Kim, Nicole Meister, Ruth Fong, and Olga
  Russakovsky.
\newblock {ELUDE: Generating interpretable explanations via a decomposition
  into labelled and unlabelled features}.
\newblock {\em arXiv}, 2022.
\newblock Decomposition in unlabelled / labelled features. Sparse explanation,
  not prediction as it uses ground truth attribute labels for explanation. Does
  not have to be faithful with actually done prediction.

\bibitem[Reinders et~al.(2022)]{chimeramix}
Christoph Reinders, Frederik Schubert, and Bodo Rosenhahn.
\newblock Chimeramix: Image classification on small datasets via masked feature
  mixing.
\newblock In Lud~De Raedt, editor, {\em Proceedings of the Thirty-First
  International Joint Conference on Artificial Intelligence, {IJCAI-22}}, pages
  1298--1305. International Joint Conferences on Artificial Intelligence
  Organization, 7 2022.
\newblock Main Track.

\bibitem[Rokade and Alluri(2021)]{rokade2021towards}
Pratyush Rokade and BKSP Kumar~Raju Alluri.
\newblock Towards quantification of explainability algorithms.
\newblock In {\em 2021 The 5th International Conference on Advances in
  Artificial Intelligence (ICAAI)}, pages 31--37, 2021.

\bibitem[Rudolph et~al.(2022)]{Rudolph_2022_WACV}
Marco Rudolph, Tom Wehrbein, Bodo Rosenhahn, and Bastian Wandt.
\newblock Fully convolutional cross-scale-flows for image-based defect
  detection.
\newblock In {\em Proceedings of the IEEE/CVF Winter Conference on Applications
  of Computer Vision (WACV)}, pages 1088--1097, January 2022.

\bibitem[R{\"u}ping et~al.(2006)]{ruping2006learning}
Stefan R{\"u}ping et~al.
\newblock Learning interpretable models.
\newblock 2006.

\bibitem[Russakovsky et~al.(2015)]{imagenet15russakovsky}
Olga Russakovsky, Jia Deng, Hao Su, Jonathan Krause, Sanjeev Satheesh, Sean Ma,
  Zhiheng Huang, Andrej Karpathy, Aditya Khosla, Michael Bernstein,
  Alexander~C. Berg, and Li Fei-Fei.
\newblock {ImageNet Large Scale Visual Recognition Challenge}.
\newblock {\em International Journal of Computer Vision (IJCV)},
  115(3):211--252, 2015.

\bibitem[Rymarczyk et~al.(2022)]{rymarczyk2022interpretable}
Dawid Rymarczyk, {\L}ukasz Struski, Micha{\l} G{\'o}rszczak, Koryna
  Lewandowska, Jacek Tabor, and Bartosz Zieli{\'n}ski.
\newblock Interpretable image classification with differentiable prototypes
  assignment.
\newblock In {\em European Conference on Computer Vision}, pages 351--368.
  Springer, 2022.

\bibitem[Rymarczyk et~al.(2021)]{rymarczyk2021protopshare}
Dawid Rymarczyk, {\L}ukasz Struski, Jacek Tabor, and Bartosz Zieli{\'n}ski.
\newblock Protopshare: Prototypical parts sharing for similarity discovery in
  interpretable image classification.
\newblock In {\em Proceedings of the 27th ACM SIGKDD Conference on Knowledge
  Discovery \& Data Mining}, pages 1420--1430, 2021.

\bibitem[Sawada and Nakamura(2022)]{10.1109/access.2022.3167702}
Yoshihide Sawada and Keigo Nakamura.
\newblock {Concept Bottleneck Model With Additional Unsupervised Concepts}.
\newblock {\em IEEE Access}, 10:41758--41765, 2022.

\bibitem[Selvaraju et~al.(2017)]{selvaraju2017grad}
Ramprasaath~R Selvaraju, Michael Cogswell, Abhishek Das, Ramakrishna Vedantam,
  Devi Parikh, and Dhruv Batra.
\newblock Grad-cam: Visual explanations from deep networks via gradient-based
  localization.
\newblock In {\em Proceedings of the IEEE international conference on computer
  vision}, pages 618--626, 2017.

\bibitem[Szegedy et~al.(2016)]{szegedy2016rethinking}
Christian Szegedy, Vincent Vanhoucke, Sergey Ioffe, Jon Shlens, and Zbigniew
  Wojna.
\newblock Rethinking the inception architecture for computer vision.
\newblock In {\em Proceedings of the IEEE conference on computer vision and
  pattern recognition}, pages 2818--2826, 2016.

\bibitem[Szegedy et~al.(2013)]{szegedy2013intriguing}
Christian Szegedy, Wojciech Zaremba, Ilya Sutskever, Joan Bruna, Dumitru Erhan,
  Ian Goodfellow, and Rob Fergus.
\newblock Intriguing properties of neural networks.
\newblock {\em arXiv preprint arXiv:1312.6199}, 2013.

\bibitem[Tao et~al.(2015)]{LDAForSelec}
Hong Tao, Chenping Hou, Feiping Nie, Yuanyuan Jiao, and Dongyun Yi.
\newblock Effective discriminative feature selection with nontrivial solution.
\newblock {\em IEEE transactions on neural networks and learning systems},
  27(4):796--808, 2015.

\bibitem[Tessera et~al.(2021)]{tessera2021keep}
Kale-ab Tessera, Sara Hooker, and Benjamin Rosman.
\newblock Keep the gradients flowing: Using gradient flow to study sparse
  network optimization.
\newblock {\em arXiv preprint arXiv:2102.01670}, 2021.

\bibitem[Van~Horn et~al.(2015)]{7298658}
Grant Van~Horn, Steve Branson, Ryan Farrell, Scott Haber, Jessie Barry, Panos
  Ipeirotis, Pietro Perona, and Serge Belongie.
\newblock Building a bird recognition app and large scale dataset with citizen
  scientists: The fine print in fine-grained dataset collection.
\newblock In {\em 2015 IEEE Conference on Computer Vision and Pattern
  Recognition (CVPR)}, pages 595--604, 2015.

\bibitem[Wah et~al.(2011)]{wah2011caltech}
Catherine Wah, Steve Branson, Peter Welinder, Pietro Perona, and Serge
  Belongie.
\newblock The caltech-ucsd birds-200-2011 dataset.
\newblock 2011.

\bibitem[Wong et~al.(2021)]{wong2021leveraging}
Eric Wong, Shibani Santurkar, and Aleksander Madry.
\newblock Leveraging sparse linear layers for debuggable deep networks.
\newblock In {\em International Conference on Machine Learning}, pages
  11205--11216. PMLR, 2021.

\bibitem[Yang et~al.(2017)]{yang2017scalable}
Hongyu Yang, Cynthia Rudin, and Margo Seltzer.
\newblock Scalable bayesian rule lists.
\newblock In {\em International conference on machine learning}, pages
  3921--3930. PMLR, 2017.

\bibitem[Yuksekgonul et~al.(2022)]{yuksekgonul2022posthoc}
Mert Yuksekgonul, Maggie Wang, and James Zou.
\newblock Post-hoc concept bottleneck models.
\newblock In {\em ICLR 2022 Workshop on PAIR{\textasciicircum}2Struct: Privacy,
  Accountability, Interpretability, Robustness, Reasoning on Structured Data},
  2022.

\bibitem[Zeiler and Fergus(2014)]{OcclusionPaper}
Matthew~D Zeiler and Rob Fergus.
\newblock Visualizing and understanding convolutional networks.
\newblock In {\em European conference on computer vision}, pages 818--833.
  Springer, 2014.

\bibitem[Zheng et~al.(2020)]{ClassUniqueCorr}
Runkai Zheng, Zhijia Yu, Yinqi Zhang, Chris Ding, Hei~Victor Cheng, and Li Liu.
\newblock {Learning Class Unique Features in Fine-Grained Visual
  Classification}.
\newblock {\em arXiv}, 2020.

\bibitem[Zou and Hastie(2005)]{ElasticNet}
Hui Zou and Trevor Hastie.
\newblock {Regularization and variable selection via the elastic net}.
\newblock {\em Journal of the Royal Statistical Society: Series B (Statistical
  Methodology)}, 67(2):301--320, 2005.

\end{thebibliography}
\bibliographystyle{ieee_fullname}
\section*{Checklist}

\begin{enumerate}

\item For all authors...
\begin{enumerate}
  \item Do the main claims made in the abstract and introduction accurately reflect the paper's contributions and scope?
    \answerYes{See Section~\ref{sec:results}, more specifically Table~\ref{table:Accuracy incmpLoss}}
  \item Did you describe the limitations of your work?
    \answerYes{See Section ~\ref{sec:limits}}
  \item Did you discuss any potential negative societal impacts of your work?
    \answerYes{See section~\ref{sec:limits}}
  \item Have you read the ethics review guidelines and ensured that your paper conforms to them?
    \answerYes{}
\end{enumerate}

\item If you are including theoretical results...
\begin{enumerate}
  \item Did you state the full set of assumptions of all theoretical results?
    \answerNA{}
        \item Did you include complete proofs of all theoretical results?
     \answerNA{}
\end{enumerate}

\item If you ran experiments...
\begin{enumerate}
  \item Did you include the code, data, and instructions needed to reproduce the main experimental results (either in the supplemental material or as a URL)?
    \answerNo{The data is missing as it is freely available. The code is not included but the method and instructions are quite detailed.}
  \item Did you specify all the training details (e.g., data splits, hyperparameters, how they were chosen)?
    \answerNo{We mentioned everything except how they were chosen.}
        \item Did you report error bars (e.g., with respect to the random seed after running experiments multiple times)?
    \answerYes{See standard deviations in Appendix}
        \item Did you include the total amount of compute and the type of resources used (e.g., type of GPUs, internal cluster, or cloud provider)?
    \answerNo{Total figure is not available}
\end{enumerate}

\item If you are using existing assets (e.g., code, data, models) or curating/releasing new assets...
\begin{enumerate}
  \item If your work uses existing assets, did you cite the creators?
    \answerYes{See Section ~\ref{app:ImpDetails}}
  \item Did you mention the license of the assets?
    \answerNA{}
  \item Did you include any new assets either in the supplemental material or as a URL?
    \answerNo{}
  \item Did you discuss whether and how consent was obtained from people whose data you're using/curating?
    \answerNA{}
  \item Did you discuss whether the data you are using/curating contains personally identifiable information or offensive content?
    \answerNA{}
\end{enumerate}

\item If you used crowdsourcing or conducted research with human subjects...
\begin{enumerate}
  \item Did you include the full text of instructions given to participants and screenshots, if applicable?
    \answerNA{}
  \item Did you describe any potential participant risks, with links to Institutional Review Board (IRB) approvals, if applicable?
    \answerNA{}
  \item Did you include the estimated hourly wage paid to participants and the total amount spent on participant compensation?
    \answerNA{}
\end{enumerate}

\end{enumerate}
\renewcommand{\bldStatement}{. Best results per column are in bold and $\pm$ indicates the standard deviation across \fivetext{} runs.}
\renewcommand{\imgbldStatement}{ for \resnet{} on \imgnetheader using the pretrained dense model.. Best results per column are in bold and $\pm$ indicates the standard deviation across \fourtext{} runs.}
\clearpage
\appendix

\section{Appendix}
    
In this appendix, we provide implementation details and standard deviations for the experiments.
Additionally, the pseudocode for \glsentrylong{glmsaga} is shown.
Finally, we present the feature visualization technique and more ablations on~\gls{customLoss}.

\section{Detailed Results}

\begin{table*}
\resizebox{\linewidth}{!}{
\centering
\begin{tabular}{c|ccccc|ccccc|ccccc|ccccc}
\toprule

\xmark &\thead{ \textbf{86.6} \\$\pm$0.4}&\thead{ 81.8 \\$\pm$0.3}&\thead{ 85.3 \\$\pm$0.2}&\thead{ 79.5 \\$\pm$0.3}&\thead{ 83.4 \\$\pm$0.2}&\thead{ 90.0 \\$\pm$0.3}&\thead{ 88.4 \\$\pm$0.3}&\thead{ 89.4 \\$\pm$0.2}&\thead{ 87.3 \\$\pm$0.4}&\thead{ 88.1 \\$\pm$0.3}&\thead{ 84.2 \\$\pm$0.1}&\thead{ 79.5 \\$\pm$0.3}&\thead{ 83.3 \\$\pm$0.1}&\thead{ 77.3 \\$\pm$0.3}&\thead{ 80.7 \\$\pm$0.2}&\thead{ 93.2 \\$\pm$0.1}&\thead{ 90.9 \\$\pm$0.2}&\thead{ 92.6 \\$\pm$0.1}&\thead{ 89.3 \\$\pm$0.3}&\thead{ 91.1 \\$\pm$0.1}\\
\cmark &\thead{ \textbf{86.6} \\$\pm$0.2}&\thead{ \textbf{84.0} \\$\pm$0.2}&\thead{ \textbf{86.5} \\$\pm$0.1}&\thead{ \textbf{81.7} \\$\pm$0.2}&\thead{ \textbf{84.0} \\$\pm$0.3}&\thead{ \textbf{91.4} \\$\pm$0.2}&\thead{ \textbf{90.7} \\$\pm$0.3}&\thead{ \textbf{91.1} \\$\pm$0.2}&\thead{ \textbf{89.8} \\$\pm$0.4}&\thead{ \textbf{90.1} \\$\pm$0.1}&\thead{ \textbf{84.4} \\$\pm$0.2}&\thead{ \textbf{81.0} \\$\pm$0.2}&\thead{ \textbf{84.0} \\$\pm$0.0}&\thead{ \textbf{79.8} \\$\pm$0.2}&\thead{ \textbf{81.7} \\$\pm$0.1}&\thead{ \textbf{93.6} \\$\pm$0.2}&\thead{ \textbf{92.1} \\$\pm$0.3}&\thead{ \textbf{93.3} \\$\pm$0.1}&\thead{ \textbf{91.1} \\$\pm$0.1}&\thead{ \textbf{92.0} \\$\pm$0.2}\\
\bottomrule
\end{tabular}
}
\caption{Accuracy\inpercent{}\tablefinisher{\gls{customLoss}} for \resnet{}\bldStatement}
\label{table:Accuracy in stdldiv}
\end{table*}

\begin{table*}
\resizebox{\linewidth}{!}{
\centering
\begin{tabular}{c|ccccc|ccccc|ccccc|ccccc}
\toprule

\densenet &\thead{ 86.3 \\$\pm$0.0}&\thead{ 76.2 \\$\pm$0.5}&\thead{ 82.9 \\$\pm$0.5}&\thead{ 75.7 \\$\pm$0.7}&\thead{ 83.1 \\$\pm$0.1}&\thead{ \textbf{91.5} \\$\pm$0.1}&\thead{ 88.2 \\$\pm$0.5}&\thead{ 89.8 \\$\pm$0.4}&\thead{ 88.1 \\$\pm$0.2}&\thead{ 90.0 \\$\pm$0.4}&\thead{ 84.1 \\$\pm$0.1}&\thead{ 72.8 \\$\pm$0.4}&\thead{ 64.6 \\$\pm$22.8}&\thead{ 71.0 \\$\pm$0.5}&\thead{ 80.5 \\$\pm$0.3}&\thead{ 93.3 \\$\pm$0.1}&\thead{ 87.3 \\$\pm$0.4}&\thead{ 91.7 \\$\pm$0.1}&\thead{ 85.8 \\$\pm$0.3}&\thead{ 91.4 \\$\pm$0.2}\\
\incv &\thead{ 82.3 \\$\pm$0.1}&\thead{ 78.0 \\$\pm$0.3}&\thead{ 80.3 \\$\pm$0.4}&\thead{ 74.0 \\$\pm$0.7}&\thead{ 78.3 \\$\pm$0.4}&\thead{ 88.9 \\$\pm$0.1}&\thead{ 87.5 \\$\pm$0.2}&\thead{ 88.1 \\$\pm$0.2}&\thead{ 85.9 \\$\pm$0.3}&\thead{ 87.4 \\$\pm$0.2}&\thead{ 79.0 \\$\pm$0.1}&\thead{ 75.8 \\$\pm$0.2}&\thead{ 77.3 \\$\pm$0.3}&\thead{ 73.1 \\$\pm$0.2}&\thead{ 76.5 \\$\pm$0.1}&\thead{ 91.5 \\$\pm$0.1}&\thead{ 88.9 \\$\pm$0.2}&\thead{ 90.3 \\$\pm$0.2}&\thead{ 86.3 \\$\pm$0.2}&\thead{ 89.4 \\$\pm$0.2}\\
\resnet &\thead{ \textbf{86.6} \\$\pm$0.2}&\thead{ \textbf{84.0} \\$\pm$0.2}&\thead{ \textbf{86.5} \\$\pm$0.1}&\thead{ \textbf{81.7} \\$\pm$0.2}&\thead{ \textbf{84.0} \\$\pm$0.3}&\thead{ 91.4 \\$\pm$0.2}&\thead{ \textbf{90.7} \\$\pm$0.3}&\thead{ \textbf{91.1} \\$\pm$0.2}&\thead{ \textbf{89.8} \\$\pm$0.4}&\thead{ \textbf{90.1} \\$\pm$0.1}&\thead{ \textbf{84.4} \\$\pm$0.2}&\thead{ \textbf{81.0} \\$\pm$0.2}&\thead{ \textbf{84.0} \\$\pm$0.0}&\thead{ \textbf{79.8} \\$\pm$0.2}&\thead{ \textbf{81.7} \\$\pm$0.1}&\thead{ \textbf{93.6} \\$\pm$0.2}&\thead{ \textbf{92.1} \\$\pm$0.3}&\thead{ \textbf{93.3} \\$\pm$0.1}&\thead{ \textbf{91.1} \\$\pm$0.1}&\thead{ \textbf{92.0} \\$\pm$0.2}\\
\bottomrule
\end{tabular}
}
\caption{Accuracy\inpercent{}\tablefinisher{backbone}\bldStatement}
\label{table:Accuracy in stdbackbone}
\end{table*}
\begin{table*}
\centering
\begin{tabular}{ccccc}
\toprule
 \multicolumn{5}{c}{\imgnetheader}\\
 \multicolumn{3}{c|}{ $\gls{nReducedFeatures}=2048$ }&\multicolumn{2}{c}{ $\gls{nReducedFeatures}=50$ }\\
Dense  & Sparse  & Finet. & Sparse  & Finet. \\\midrule
76.1&\thead{ \imgsunl \\$\pm$0.0}&\thead{{\imgfunl} \\$\pm$0.0}&\thead{ \imgsl \\$\pm$0.1}&\thead{ \imgfl \\$\pm$0.1}\\
\bottomrule
\end{tabular}
\caption{Accuracy\inpercent{}\imgbldStatement}
\label{table:Accuracy imgnet}
\end{table*}

The full results of Section~\ref{sec:results} with the standard deviations are presented in
\suppt
Tables~\ref{table:Accuracy in stdldiv} to~\ref{table:loc5 incmpLoss}. 
The reported standard deviations are, except for \densenet{} on \birdsheader{}, as mentioned in Section~\ref{sec:results},  generally rather small compared to the differences in means, which supports our conclusions. 
We also show exemplary images with the 5 most important features for the dense and sparse conventional model in Figures~\ref{app:fig:vizExamples0} to~\ref{app:fig:vizExamples8}. The comparison with the finetuned~\gls{layerName} shows an improved localization and interpretability of the features.
The finetuned \gls{layerName} in Figure~\ref{app:fig:vizExamples3} seems to use features that each individually focus more on chest, lower belly, head, bill or crown, whereas for the dense and sparse models the different features focus on the same regions. This increased \loc{5} and with it interpretability was also measured in Section~\ref{sec:results}.
\subsection{Implementation Details}
\label{app:ImpDetails}
We use Pytorch~\citep{Pytorch} to implement our methods and on ImageNet pretrained models 
as backbone feature extractor. We utilized \glm{} and \textit{robustness}~\citep{robustness}.
The images are resized to $448\times448$ ($299\times299$ for \incv, $224\times224$ for \imgnetheader), normalized, randomly horizontally flipped and jitter is applied.
The model is finetuned using stochastic gradient descent on the specific dataset for $150$ ($100$ for \birdsheader) epochs with a batch size of $16$ ($64$ for \gls{imgnetheader}), starting with $5\cdot10^{-3}$ as learning rate for the pretrained layer and $0.01$ for the final linear layer. 
Both get multiplied by $0.4$ every $30$ epochs. Additionally, we used momentum of $0.9$, $\ell_2$-regularization of $5\cdot10^{-4}$ and apply a dropout rate of $0.2$ on the features to reduce dependencies. \gls{cLW} was set to $0.196$ for \resnet{}, $0.098$ for \densenet{} and $0.049$ for \incv.
For the feature selection, we set $\gls{elaW} = 0.8$ and reduce the regularization strength \gls{elaWeight} by $90\,\%$ as we found it sped up the process without decreasing performance.\\
We use \glm{} to compute the regularization path with $\gls{elaW}=0.99$ and all other parameters set to default with a lookbehind of $T=5$.
From this path, the solution with maximum $\gls{nperClass} \leq 10$ is selected. 
Then the non-zero entries with the lowest absolute value get zeroed out until we are left with $\gls{nperClass} = 5$,
as we empirically found that they do not improve test accuracy after finetuning.\\
This selected solution replaces the final layer of our model. 
Then we train for $40$ epochs, starting with the final learning rate of the initial training multiplied by $100$ ($\frac{1}{100}$ of that for \imgnetheader), and decrease it by $60\,\%$ every $10$ epochs. 
Dropout on the features was set to $0.1$ and momentum was increased to $0.95$. Note that, while the increased momentum has been important for the stability of the final training, the hyperparameters were not thoroughly optimized for the sparse case. 
\subsubsection{Competitors}
\label{app:cbmjoint}
For creating the accuracy for \resnet{} and \textit{\glsentrylong{cbm} - joint} in Table~\ref{tab:competitors} we resized the images to $448\times448$ and used a batch size of $16$. The remaining used hyperparameters were almost identical to the \gls{cbm} experiments with \incv, but we only trained for up to $400$ epochs, as $650$ led to decreased accuracy ($-0.8$ percent points). 
Additionally, the learning rate was not decayed, mirroring the published code.
The reported accuracy stems from three runs with a standard deviation of $0.7$.\\
For MCL, we used the reported hyperparameters of $\mu=0.005$ and $\lambda=10$. For finetuning, we assigned every feature to every class that was using it. We optimized the hyperparameters for FRL based on accuracy, leading to $K = 10$ and $\lambda=0.01$.\\ 

\section{\Gls{glmsaga}}
\label{app:glmsaga}
\begin{algorithm}[!t]
	\caption{Pseudocode from \glsentrylong{glmsaga}
	}
	\label{alg:solver}
	\begin{algorithmic}[1]
		\STATE Initialize table of scalars $a_i' = 0$ for $i \in [n]$
		\STATE Initialize average gradient of table $g_{avg}=0$ and 
		$g_{0avg}=0$
		\FOR{minibatch $B\subset [n]$}
		\FOR{$i \in B$}
		\STATE $a_i = x_i^T\beta + \beta_0 - y_i$
		\STATE $g_i = a_i \cdot x_i$ \textit{// calculate new gradient information}
		\STATE $g_i' = a_i' \cdot x_i$ \textit{// calculate stored gradient 
			information}
		\ENDFOR
		\STATE $g = \frac{1}{|B|}\sum_{i \in B} g_i$
		\STATE $g' = \frac{1}{|B|}\sum_{i \in B} g_i'$
		
		\STATE $g_0 = \frac{1}{|B|}\sum_{i \in B} a_i$
		\STATE $g_0' = \frac{1}{|B|}\sum_{i \in B} a_i'$
		
		\STATE $\beta = \beta - \gamma(g - g' + g_{avg})$
		\STATE $\beta_0 = \beta_0 - \gamma(g_0 - g_0' + g_{0avg})$
		\STATE $\beta = \textrm{Prox}_{\gamma\lambda\alpha, 
			\gamma\lambda(1-\alpha)}(\beta)$
		\FOR{$i\in B$}
		\STATE $a_i' = a_i$ \textit{// update table}
		\STATE $g_{avg} = g_{avg} + \frac{|B|}{n}(g - g')$ \textit{// update 
			average}
		\STATE $g_{0avg} = g_{0avg} + \frac{|B|}{n}(g_0 - g_0')$ 
		\ENDFOR
		\ENDFOR
	\end{algorithmic}
\end{algorithm}
This section includes the Pseudocode for \glsentrylong{glmsaga} in algorithm~\ref{alg:solver}. The proximal operator $\textrm{Prox}_{\lambda_1, \lambda_2}(\beta)$ is defined as:
\begin{equation}
\textrm{Prox}_{\lambda_1, \lambda_2}(\beta) = \begin{cases}
\frac{\beta - \lambda_1}{1+\lambda_2} &\text{if } \beta > \lambda_1 \\
\frac{\beta + \lambda_1}{1+\lambda_2} &\text{if } \beta < \lambda_1 \\
0 &\text{otherwise}
\end{cases}
\end{equation}
\section{Visualization of Features}
\label{app:featureViz}
For feature visualization, we follow a masking approach. We systematically blur, following~\citep{fong2017interpretable}, one patch of size $p\times p$ of the image and measure the difference in feature activation between the augmented image and not augmented image. The actual localization map $\glsentrylong{LocalizationMaps}$ for that square size is computed by 
\begin{align}
    \gls{LocalizationMaps}_{pxy} =  \text{ReLU}(\glsentrylong{featureVector}(I) - \glsentrylong{featureVector}(I_{pxy}))
\end{align}
where $I_{pxy}$ indicates the image where a $p$-sized patch starting at position $(x *p,y*p)$ is blurred and the ReLU suppresses parts that increased the feature activation, since blur should not be injecting a feature. The final localization map is the combination of different square sizes $p\in\{28,56,64,112,224\}$ to accommodate for differently sized features:
\begin{equation}
    \gls{LocalizationMaps}= \sum_{p}\frac{\gls{LocalizationMaps}_p}{\max(\gls{LocalizationMaps}_p)}
\end{equation}
Notably, $\gls{LocalizationMaps}_p$ has to be resized according to the smallest $p$ and we only show $\gls{LocalizationMaps}^i$ for one feature $i$.
\section{Feature Alignment}
\label{app:sec:alignment}
In this section, we use the alignment of the shown feature in Figure~\ref{fig:4Strahlig}  with four-engine aircraft to exemplary show how one can align a feature manually. We first visualize the distribution of the feature over the training data in Figure~\ref{fig:distair}. This indicates a more binary attribute and by investigating the images and saliency maps, we observed an alignment with four-engine aircraft. 
We test the hypothesis by filtering for the classes "A340-500" and "BAE 146-300", for which the feature corresponds to the four engines. 
Figure~\ref{fig:lowestair} shows that the lowest activating examples of this group do not clearly show the four-engines which supports our hypothesis.
Note that one feature can correspond to multiple concepts, as another hypothesis is an alignment with the propeller.
Whether all correlating concepts have to be understood or how exact the analysis has to be depends on the application.
\begin{figure*}[t]
\begin{center}
  \includegraphics[width=\linewidth]{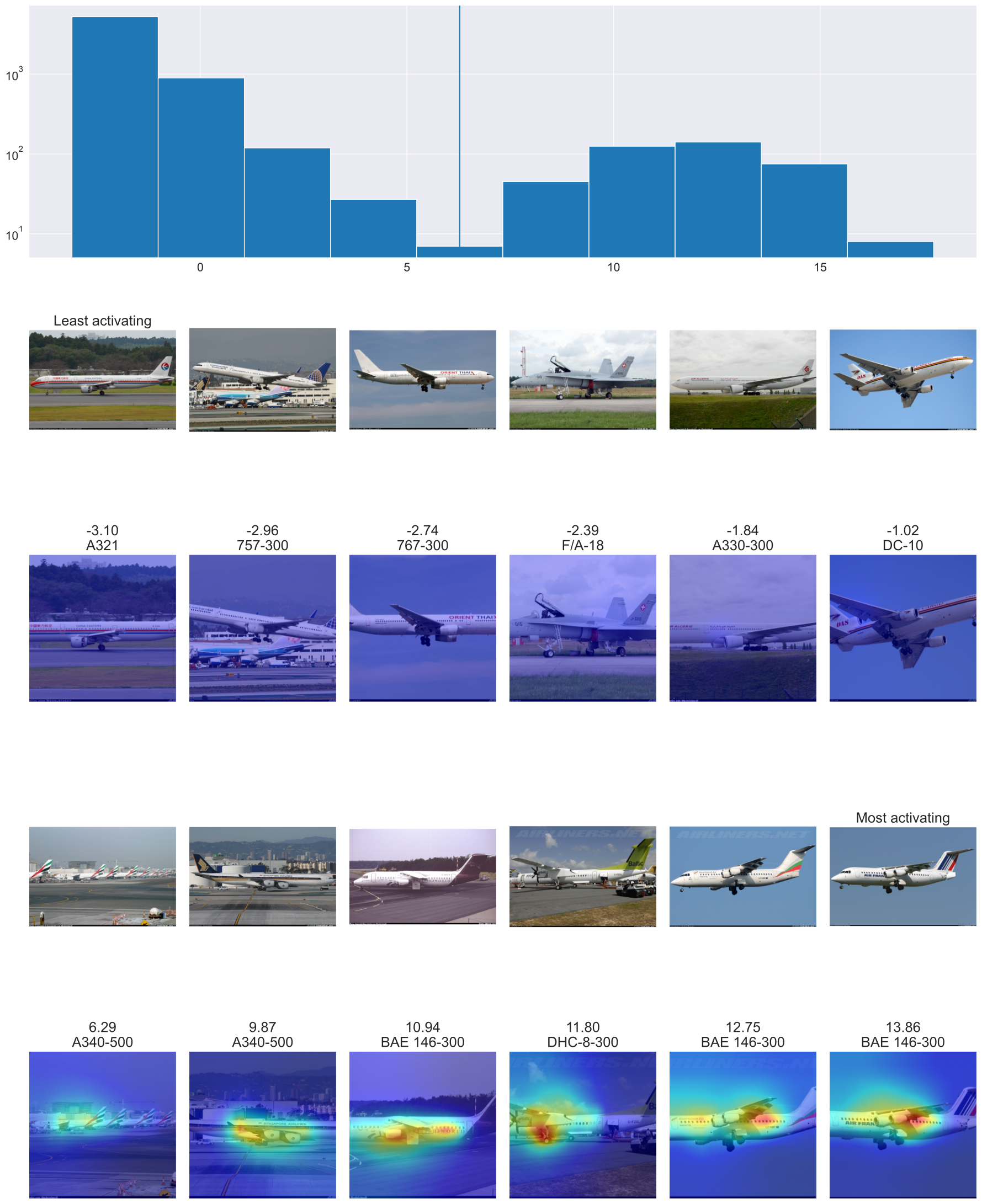}
\end{center}
  \caption{Top: Distribution of feature activation over the training data. Bottom: Different examples of this distribution with their scaled feature localization. The activation and class name is given above the image. }
   
\vspace{-0.55cm}
\label{fig:distair}
\end{figure*}
\begin{figure*}
     \begin{subfigure}[t]{.3\textwidth}
         \centering
         \includegraphics[width=\textwidth]{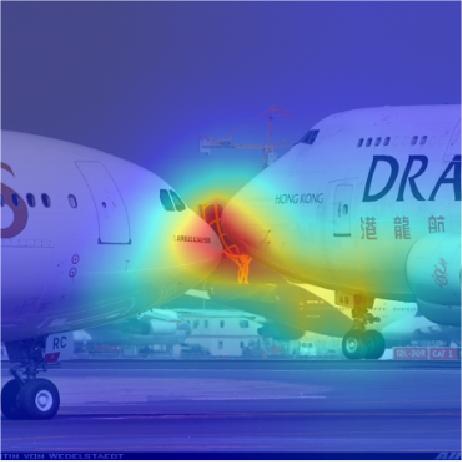}
     \end{subfigure}
     \hfill
     \begin{subfigure}[t]{.3\textwidth}
         \centering
       \includegraphics[width=\textwidth]{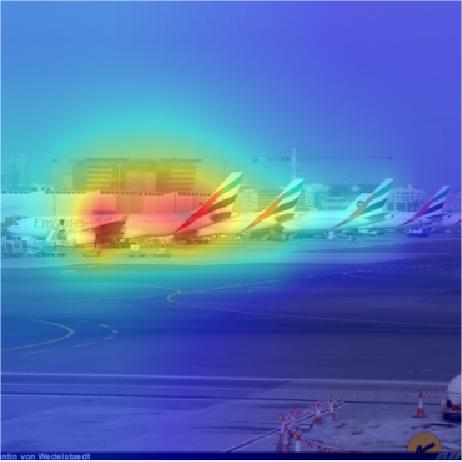}
     \end{subfigure}
     \hfill
     \begin{subfigure}[t]{.3\textwidth}
         \centering
       \includegraphics[width=\textwidth]{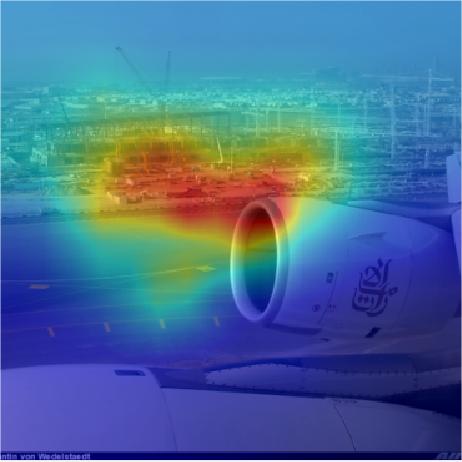}
     \end{subfigure}
        \caption{The three least activating training examples of classes "A340-500" and "BAE-16-300" for the given feature. }
        \label{fig:lowestair}
\end{figure*}
\section{Ablations on Feature Diversity Loss}
\label{app:AblCustom}
In this section, we present an additional analysis of the factors in~\gls{customLoss} and the impact of \gls{cLW}. 
\begin{table*}
\resizebox{\linewidth}{!}{
\centering
\begin{tabular}{c|ccccc|ccccc|ccccc|ccccc}
\toprule

\xmark &\thead{ 50.2 \\$\pm$0.2}&\thead{ 46.0 \\$\pm$0.2}&\thead{ 43.4 \\$\pm$0.3}&\thead{ 48.0 \\$\pm$0.5}&\thead{ 46.5 \\$\pm$0.3}&\thead{ 46.5 \\$\pm$0.5}&\thead{ 43.8 \\$\pm$0.7}&\thead{ 40.9 \\$\pm$0.4}&\thead{ 45.9 \\$\pm$0.6}&\thead{ 44.0 \\$\pm$0.6}&\thead{ 41.5 \\$\pm$0.2}&\thead{ 38.4 \\$\pm$0.1}&\thead{ 34.9 \\$\pm$0.1}&\thead{ 40.8 \\$\pm$0.6}&\thead{ 35.1 \\$\pm$0.7}&\thead{ 45.0 \\$\pm$0.2}&\thead{ 41.7 \\$\pm$0.3}&\thead{ 39.1 \\$\pm$0.1}&\thead{ 43.6 \\$\pm$0.6}&\thead{ 43.7 \\$\pm$0.4}\\
\cmark &\thead{ \textbf{98.9} \\$\pm$0.1}&\thead{ \textbf{69.9} \\$\pm$0.5}&\thead{ \textbf{71.9} \\$\pm$0.4}&\thead{ \textbf{65.2} \\$\pm$1.4}&\thead{ \textbf{72.6} \\$\pm$0.3}&\thead{ \textbf{98.8} \\$\pm$0.2}&\thead{ \textbf{85.7} \\$\pm$1.5}&\thead{ \textbf{86.6} \\$\pm$1.4}&\thead{ \textbf{69.3} \\$\pm$0.8}&\thead{ \textbf{73.9} \\$\pm$1.3}&\thead{ \textbf{98.7} \\$\pm$0.1}&\thead{ \textbf{69.5} \\$\pm$1.1}&\thead{ \textbf{81.0} \\$\pm$1.2}&\thead{ \textbf{70.7} \\$\pm$1.9}&\thead{ \textbf{85.3} \\$\pm$1.0}&\thead{ \textbf{99.2} \\$\pm$0.1}&\thead{ \textbf{72.4} \\$\pm$2.0}&\thead{ \textbf{74.6} \\$\pm$1.5}&\thead{ \textbf{63.7} \\$\pm$1.3}&\thead{ \textbf{74.8} \\$\pm$0.9}\\
\bottomrule
\end{tabular}
}
\caption{\loc{5}\inpercent{}\tablefinisher{\gls{customLoss}} for \resnet{}\bldStatement}
\label{table:loc5 in stdldiv}
\end{table*}

\begin{table*}
\resizebox{\linewidth}{!}{
\centering
\begin{tabular}{c|ccccc|ccccc|ccccc|ccccc}
\toprule

\densenet &\thead{ 98.5 \\$\pm$0.1}&\thead{ 69.1 \\$\pm$1.7}&\thead{ 64.2 \\$\pm$1.2}&\thead{ \textbf{76.2} \\$\pm$1.0}&\thead{ 71.2 \\$\pm$0.8}&\thead{ \textbf{99.0} \\$\pm$0.1}&\thead{ 40.4 \\$\pm$0.7}&\thead{ 39.9 \\$\pm$0.9}&\thead{ 64.3 \\$\pm$2.0}&\thead{ 62.6 \\$\pm$2.1}&\thead{ 98.6 \\$\pm$0.1}&\thead{ 45.7 \\$\pm$1.1}&\thead{ 42.3 \\$\pm$3.5}&\thead{ 63.1 \\$\pm$3.1}&\thead{ 66.2 \\$\pm$2.2}&\thead{ 98.7 \\$\pm$0.1}&\thead{ 47.4 \\$\pm$1.3}&\thead{ 46.3 \\$\pm$1.0}&\thead{ \textbf{71.8} \\$\pm$1.5}&\thead{ 68.0 \\$\pm$1.2}\\
\incv &\thead{ 86.3 \\$\pm$0.5}&\thead{ \textbf{74.9} \\$\pm$0.9}&\thead{ 65.1 \\$\pm$0.6}&\thead{ 53.2 \\$\pm$0.8}&\thead{ 52.7 \\$\pm$0.4}&\thead{ 95.1 \\$\pm$0.2}&\thead{ \textbf{87.2} \\$\pm$1.6}&\thead{ 72.2 \\$\pm$1.4}&\thead{ 53.7 \\$\pm$1.1}&\thead{ 56.1 \\$\pm$0.9}&\thead{ 81.8 \\$\pm$1.0}&\thead{ 47.9 \\$\pm$1.0}&\thead{ 47.0 \\$\pm$0.8}&\thead{ 42.7 \\$\pm$1.6}&\thead{ 45.6 \\$\pm$1.0}&\thead{ 92.0 \\$\pm$0.3}&\thead{ \textbf{76.3} \\$\pm$0.6}&\thead{ 63.7 \\$\pm$0.6}&\thead{ 52.0 \\$\pm$1.2}&\thead{ 50.7 \\$\pm$0.8}\\
\resnet &\thead{ \textbf{98.9} \\$\pm$0.1}&\thead{ 69.9 \\$\pm$0.5}&\thead{ \textbf{71.9} \\$\pm$0.4}&\thead{ 65.2 \\$\pm$1.4}&\thead{ \textbf{72.6} \\$\pm$0.3}&\thead{ 98.8 \\$\pm$0.2}&\thead{ 85.7 \\$\pm$1.5}&\thead{ \textbf{86.6} \\$\pm$1.4}&\thead{ \textbf{69.3} \\$\pm$0.8}&\thead{ \textbf{73.9} \\$\pm$1.3}&\thead{ \textbf{98.7} \\$\pm$0.1}&\thead{ \textbf{69.5} \\$\pm$1.1}&\thead{ \textbf{81.0} \\$\pm$1.2}&\thead{ \textbf{70.7} \\$\pm$1.9}&\thead{ \textbf{85.3} \\$\pm$1.0}&\thead{ \textbf{99.2} \\$\pm$0.1}&\thead{ 72.4 \\$\pm$2.0}&\thead{ \textbf{74.6} \\$\pm$1.5}&\thead{ 63.7 \\$\pm$1.3}&\thead{ \textbf{74.8} \\$\pm$0.9}\\
\bottomrule
\end{tabular}
}
\caption{\loc{5}\inpercent{}\tablefinisher{backbone}\bldStatement}
\label{table:loc5 in stdbackbone}
\end{table*}

\begin{table*}
\resizebox{\linewidth}{!}{
\centering
\begin{tabular}{c|ccccc|ccccc|ccccc}
\toprule

\xmark &\thead{ \textbf{86.6} \\$\pm$0.4}&\thead{ 81.8 \\$\pm$0.3}&\thead{ 85.3 \\$\pm$0.2}&\thead{ 79.5 \\$\pm$0.3}&\thead{ 83.4 \\$\pm$0.2}&\thead{ 90.0 \\$\pm$0.3}&\thead{ 88.4 \\$\pm$0.3}&\thead{ 89.4 \\$\pm$0.2}&\thead{ 87.3 \\$\pm$0.4}&\thead{ 88.1 \\$\pm$0.3}&\thead{ 93.2 \\$\pm$0.1}&\thead{ 90.9 \\$\pm$0.2}&\thead{ 92.6 \\$\pm$0.1}&\thead{ 89.3 \\$\pm$0.3}&\thead{ 91.1 \\$\pm$0.1}\\
\cmark &\thead{ \textbf{86.6} \\$\pm$0.2}&\thead{ \textbf{84.0} \\$\pm$0.2}&\thead{ \textbf{86.5} \\$\pm$0.1}&\thead{ \textbf{81.7} \\$\pm$0.2}&\thead{ \textbf{84.0} \\$\pm$0.3}&\thead{ \textbf{91.4} \\$\pm$0.2}&\thead{ \textbf{90.7} \\$\pm$0.3}&\thead{ \textbf{91.1} \\$\pm$0.2}&\thead{ \textbf{89.8} \\$\pm$0.4}&\thead{ \textbf{90.1} \\$\pm$0.1}&\thead{ \textbf{93.6} \\$\pm$0.2}&\thead{ \textbf{92.1} \\$\pm$0.3}&\thead{ \textbf{93.3} \\$\pm$0.1}&\thead{ \textbf{91.1} \\$\pm$0.1}&\thead{ \textbf{92.0} \\$\pm$0.2}\\
\bottomrule MCL~\citep{DevilChannels} & \thead{86.1\\$\pm$0.2} & \thead{81.9\\$\pm$0.3} & \thead{85.1\\$\pm$0.2} & \thead{79.4\\$\pm$0.2} & \thead{82.8\\$\pm$0.1} & \thead{90.1\\$\pm$0.1} & \thead{88.4\\$\pm$0.2} & \thead{89.0\\$\pm$0.2} & \thead{87.2\\$\pm$0.5} & \thead{88.1\\$\pm$0.1} & \thead{93.1\\$\pm$0.1} & \thead{91.0\\$\pm$0.2} & \thead{92.5\\$\pm$0.2} & \thead{89.0\\$\pm$0.5} & \thead{90.7\\$\pm$0.3}\\FRL~\citep{ClassUniqueCorr} & \thead{86.4\\$\pm$0.2} & \thead{81.5\\$\pm$0.2} & \thead{85.3\\$\pm$0.2} & \thead{78.9\\$\pm$0.5} & \thead{82.6\\$\pm$0.5} & \thead{90.0\\$\pm$0.1} & \thead{88.5\\$\pm$0.2} & \thead{89.4\\$\pm$0.4} & \thead{87.5\\$\pm$0.2} & \thead{88.2\\$\pm$0.2} & \thead{93.3\\$\pm$0.1} & \thead{90.8\\$\pm$0.3} & \thead{92.6\\$\pm$0.2} & \thead{89.4\\$\pm$0.2} & \thead{90.9\\$\pm$0.2}\\
\end{tabular}
}
\caption{Accuracy\inpercent{} for \resnet{} \cmpLoss}
\label{table:Accuracy incmpLoss}
\end{table*}

\begin{table*}
\resizebox{\linewidth}{!}{
\centering
\begin{tabular}{c|ccccc|ccccc|ccccc}
\toprule

\xmark &\thead{ 50.2 \\$\pm$0.2}&\thead{ 46.0 \\$\pm$0.2}&\thead{ 43.4 \\$\pm$0.3}&\thead{ 48.0 \\$\pm$0.5}&\thead{ 46.5 \\$\pm$0.3}&\thead{ 46.5 \\$\pm$0.5}&\thead{ 43.8 \\$\pm$0.7}&\thead{ 40.9 \\$\pm$0.4}&\thead{ 45.9 \\$\pm$0.6}&\thead{ 44.0 \\$\pm$0.6}&\thead{ 45.0 \\$\pm$0.2}&\thead{ 41.7 \\$\pm$0.3}&\thead{ 39.1 \\$\pm$0.1}&\thead{ 43.6 \\$\pm$0.6}&\thead{ 43.7 \\$\pm$0.4}\\
\cmark &\thead{ \textbf{98.9} \\$\pm$0.1}&\thead{ \textbf{69.9} \\$\pm$0.5}&\thead{ \textbf{71.9} \\$\pm$0.4}&\thead{ \textbf{65.2} \\$\pm$1.4}&\thead{ \textbf{72.6} \\$\pm$0.3}&\thead{ \textbf{98.8} \\$\pm$0.2}&\thead{ \textbf{85.7} \\$\pm$1.5}&\thead{ \textbf{86.6} \\$\pm$1.4}&\thead{ \textbf{69.3} \\$\pm$0.8}&\thead{ \textbf{73.9} \\$\pm$1.3}&\thead{ \textbf{99.2} \\$\pm$0.1}&\thead{ \textbf{72.4} \\$\pm$2.0}&\thead{ \textbf{74.6} \\$\pm$1.5}&\thead{ \textbf{63.7} \\$\pm$1.3}&\thead{ \textbf{74.8} \\$\pm$0.9}\\
\bottomrule MCL~\citep{DevilChannels} & \thead{52.5\\$\pm$0.3} & \thead{51.4\\$\pm$0.7} & \thead{48.9\\$\pm$0.8} & \thead{56.7\\$\pm$1.0} & \thead{52.3\\$\pm$1.0} & \thead{50.1\\$\pm$0.7} & \thead{50.6\\$\pm$0.8} & \thead{48.3\\$\pm$1.2} & \thead{51.7\\$\pm$1.8} & \thead{50.1\\$\pm$1.9} & \thead{49.0\\$\pm$0.6} & \thead{49.0\\$\pm$0.9} & \thead{46.2\\$\pm$0.8} & \thead{51.8\\$\pm$1.2} & \thead{49.5\\$\pm$1.3}\\FRL~\citep{ClassUniqueCorr} & \thead{51.1\\$\pm$0.3} & \thead{47.1\\$\pm$0.5} & \thead{44.1\\$\pm$0.4} & \thead{49.0\\$\pm$0.7} & \thead{46.3\\$\pm$0.6} & \thead{48.2\\$\pm$0.3} & \thead{44.6\\$\pm$0.3} & \thead{41.2\\$\pm$0.4} & \thead{44.9\\$\pm$0.8} & \thead{43.1\\$\pm$0.5} & \thead{46.2\\$\pm$0.1} & \thead{43.0\\$\pm$0.4} & \thead{40.0\\$\pm$0.4} & \thead{43.0\\$\pm$1.1} & \thead{41.7\\$\pm$1.1}\\
\end{tabular}
}
\caption{\loc{5}\inpercent{} for \resnet{} \cmpLoss}
\label{table:loc5 incmpLoss}
\end{table*}

\subsection{Factor Importance}
\begin{table*}
\resizebox{\linewidth}{!}{
\centering
\begin{tabular}{c|ccccc|ccccc|ccccc}
\toprule
&\multicolumn{5}{c|}{CUB-2011}&\multicolumn{5}{c|}{FGVC-Aircraft}&\multicolumn{5}{c}{Stanford Cars}\\
 &\multicolumn{3}{c|}{$\gls{nReducedFeatures}=2048$}&\multicolumn{2}{c|}{$\gls{nReducedFeatures}=50$}& \multicolumn{3}{c|}{$\gls{nReducedFeatures}=2048$}&\multicolumn{2}{c|}{$\gls{nReducedFeatures}=50$} &
 \multicolumn{3}{c}{$\gls{nReducedFeatures}=2048$} & \multicolumn{2}{c}{$\gls{nReducedFeatures}=50$}\\
 Loss & Dense  & Sparse  & Finet. & Sparse  & Finet. & Dense  & Sparse  & Finet. & Sparse  & Finet. & Dense  & Sparse  & Finet. & Sparse  & Finet. \\\midrule
\gls{customLoss} &\thead{ \textbf{86.6} \\$\pm$0.2}&\thead{ \textbf{84.0} \\$\pm$0.2}&\thead{ \textbf{86.5} \\$\pm$0.1}&\thead{ \textbf{81.7} \\$\pm$0.2}&\thead{ \textbf{84.0} \\$\pm$0.3}&\thead{ \textbf{91.4} \\$\pm$0.2}&\thead{ \textbf{90.7} \\$\pm$0.3}&\thead{ \textbf{91.1} \\$\pm$0.2}&\thead{ \textbf{89.8} \\$\pm$0.4}&\thead{ \textbf{90.1} \\$\pm$0.1}&\thead{ \textbf{93.6} \\$\pm$0.2}&\thead{ \textbf{92.1} \\$\pm$0.3}&\thead{ \textbf{93.3} \\$\pm$0.1}&\thead{ \textbf{91.1} \\$\pm$0.1}&\thead{ \textbf{92.0} \\$\pm$0.2}\\
\SoftmaxName &\thead{ 86.3 \\$\pm$0.2}&\thead{ 82.7 \\$\pm$0.4}&\thead{ 85.6 \\$\pm$0.3}&\thead{ 80.2 \\$\pm$0.5}&\thead{ 83.4 \\$\pm$0.3}&\thead{ 90.8 \\$\pm$0.4}&\thead{ 89.7 \\$\pm$0.3}&\thead{ 90.1 \\$\pm$0.2}&\thead{ 88.7 \\$\pm$0.4}&\thead{ 89.1 \\$\pm$0.3}&\thead{ 93.2 \\$\pm$0.1}&\thead{ 91.0 \\$\pm$0.2}&\thead{ 92.8 \\$\pm$0.2}&\thead{ 89.9 \\$\pm$0.1}&\thead{ 91.5 \\$\pm$0.2}\\
\classWeightsName &\thead{ 86.3 \\$\pm$0.2}&\thead{ 82.0 \\$\pm$0.2}&\thead{ 85.4 \\$\pm$0.3}&\thead{ 79.0 \\$\pm$0.2}&\thead{ 83.1 \\$\pm$0.2}&\thead{ 90.2 \\$\pm$0.3}&\thead{ 88.6 \\$\pm$0.4}&\thead{ 89.6 \\$\pm$0.3}&\thead{ 87.3 \\$\pm$0.2}&\thead{ 88.5 \\$\pm$0.4}&\thead{ 93.2 \\$\pm$0.2}&\thead{ 90.8 \\$\pm$0.2}&\thead{ 92.6 \\$\pm$0.2}&\thead{ 89.1 \\$\pm$0.3}&\thead{ 91.0 \\$\pm$0.3}\\
\bottomrule
\end{tabular}
}
\caption{Impact of factors in~\gls{customLoss} on accuracy\inpercent{} for \resnet{}\bldStatement}
\label{table:Impact of factors in customLoss on Impact of factors in customLoss on Accuracy in std}
\end{table*}
\begin{table*}
\resizebox{\linewidth}{!}{
\centering
\begin{tabular}{c|ccccc|ccccc|ccccc}
\toprule

\gls{customLoss} &\thead{ 98.9 \\$\pm$0.1}&\thead{ \textbf{69.9} \\$\pm$0.5}&\thead{ \textbf{71.9} \\$\pm$0.4}&\thead{ 65.2 \\$\pm$1.4}&\thead{ 72.6 \\$\pm$0.3}&\thead{ 98.8 \\$\pm$0.2}&\thead{ \textbf{85.7} \\$\pm$1.5}&\thead{ \textbf{86.6} \\$\pm$1.4}&\thead{ \textbf{69.3} \\$\pm$0.8}&\thead{ \textbf{73.9} \\$\pm$1.3}&\thead{ 99.2 \\$\pm$0.1}&\thead{ \textbf{72.4} \\$\pm$2.0}&\thead{ \textbf{74.6} \\$\pm$1.5}&\thead{ 63.7 \\$\pm$1.3}&\thead{ \textbf{74.8} \\$\pm$0.9}\\
\SoftmaxName &\thead{ \textbf{99.3} \\$\pm$0.0}&\thead{ 67.6 \\$\pm$0.9}&\thead{ 66.9 \\$\pm$0.5}&\thead{ \textbf{70.5} \\$\pm$2.0}&\thead{ \textbf{73.3} \\$\pm$1.8}&\thead{ \textbf{99.7} \\$\pm$0.0}&\thead{ 72.4 \\$\pm$1.2}&\thead{ 74.6 \\$\pm$1.6}&\thead{ 68.5 \\$\pm$4.3}&\thead{ 73.6 \\$\pm$2.1}&\thead{ \textbf{99.7} \\$\pm$0.0}&\thead{ 64.9 \\$\pm$0.8}&\thead{ 65.3 \\$\pm$0.6}&\thead{ \textbf{68.3} \\$\pm$3.6}&\thead{ 73.5 \\$\pm$2.0}\\
\classWeightsName &\thead{ 50.6 \\$\pm$0.3}&\thead{ 46.2 \\$\pm$0.3}&\thead{ 43.5 \\$\pm$0.2}&\thead{ 48.1 \\$\pm$1.1}&\thead{ 46.8 \\$\pm$0.6}&\thead{ 47.6 \\$\pm$0.6}&\thead{ 44.3 \\$\pm$0.5}&\thead{ 41.2 \\$\pm$0.3}&\thead{ 45.0 \\$\pm$1.7}&\thead{ 43.7 \\$\pm$1.6}&\thead{ 45.5 \\$\pm$0.2}&\thead{ 42.2 \\$\pm$0.3}&\thead{ 39.5 \\$\pm$0.2}&\thead{ 42.7 \\$\pm$1.1}&\thead{ 42.5 \\$\pm$1.1}\\
\bottomrule
\end{tabular}
}
\caption{Impact of factors in~\gls{customLoss} on \loc{5}\inpercent{} for \resnet{}\bldStatement}
\label{table:Impact of factors in customLoss on Impact of factors in customLoss on loc5 in std}
\end{table*}

We analyzed the impact of the two factors in Equation~\ref{eq:ScaleDiv} with for accuracy optimized \gls{cLW}, shown in Tables~\ref{table:Impact of factors in customLoss on Impact of factors in customLoss on Accuracy in std} and~\ref{table:Impact of factors in customLoss on Impact of factors in customLoss on loc5 in std}. 
The label \classWeightsName{} indicates not using the weights of the predicted class and \SoftmaxName{} refers to not maintaining their relative mean. 
We used $\gls{cLW}= 0.001\cdot\frac{196}{2048}\approx1\mathrm{e}{-5}$ for
\classWeightsName{}, since we use the size of the feature maps with $196=\gls{featuresMapwidth}\cdot\gls{featuresMapheigth}$ and the number of features of the
baseline model$~\gls{nFeatures}=2048$ as scaling factors in order to be less dependent of model architecture and image size, and $\gls{cLW}= 0.1$ for \SoftmaxName.
Only the combination of both factors leads to an improved accuracy, validating our idea that it is important to only enforce diversity of features that are found in the input and used in conjunction.

\subsection{Loss Weighting}
This section is concerned with the impact of the weighting factor~\gls{cLW} for the feature diversity loss~\gls{customLoss}. For our proposed method, we use $\gls{cLW}=0.196$.
Tables~\ref{table:Accuracy in stdbeta} and~\ref{table:loc5 in stdbeta} show that
\gls{customLoss} improves the \loc{5} and accuracy 
across all datasets
in the sparse case with increasing~\gls{cLW} up to a maximum roughly around $\gls{cLW}=1$. Setting the value higher leads to~\gls{customLoss} dominating the training. To ensure that the network is still mainly optimized for classification, we choose $\gls{cLW}=0.196$, even though in some cases we could still observe a slight gain in accuracy with a slight increase of~\gls{cLW}.
The positive relation between~\loc{5} and accuracy, visualized in
\suppt
Figure~\ref{fig:DivAcc}, supports our approach of enforcing varied features for the extremely sparse case.
\begin{table*}
\resizebox{\linewidth}{!}{
\centering
\begin{tabular}{c|ccccc|ccccc|ccccc}
\toprule
&\multicolumn{5}{c|}{CUB-2011}&\multicolumn{5}{c|}{FGVC-Aircraft}&\multicolumn{5}{c}{Stanford Cars}\\
 &\multicolumn{3}{c|}{$\gls{nReducedFeatures}=2048$}&\multicolumn{2}{c|}{$\gls{nReducedFeatures}=50$}& \multicolumn{3}{c|}{$\gls{nReducedFeatures}=2048$}&\multicolumn{2}{c|}{$\gls{nReducedFeatures}=50$} &
 \multicolumn{3}{c}{$\gls{nReducedFeatures}=2048$} & \multicolumn{2}{c}{$\gls{nReducedFeatures}=50$}\\
 \gls{cLW} & Dense  & Sparse  & Finet. & Sparse  & Finet. & Dense  & Sparse  & Finet. & Sparse  & Finet. & Dense  & Sparse  & Finet. & Sparse  & Finet. \\\midrule
0 &\thead{ \textbf{86.6} \\$\pm$0.4}&\thead{ 81.8 \\$\pm$0.3}&\thead{ 85.3 \\$\pm$0.2}&\thead{ 79.5 \\$\pm$0.3}&\thead{ 83.4 \\$\pm$0.2}&\thead{ 90.0 \\$\pm$0.3}&\thead{ 88.4 \\$\pm$0.3}&\thead{ 89.4 \\$\pm$0.2}&\thead{ 87.3 \\$\pm$0.4}&\thead{ 88.1 \\$\pm$0.3}&\thead{ 93.2 \\$\pm$0.1}&\thead{ 90.9 \\$\pm$0.2}&\thead{ 92.6 \\$\pm$0.1}&\thead{ 89.3 \\$\pm$0.3}&\thead{ 91.1 \\$\pm$0.1}\\
0.00196 &\thead{ 86.4 \\$\pm$0.2}&\thead{ 82.0 \\$\pm$0.3}&\thead{ 85.5 \\$\pm$0.3}&\thead{ 79.3 \\$\pm$0.3}&\thead{ 83.3 \\$\pm$0.2}&\thead{ 90.2 \\$\pm$0.2}&\thead{ 88.7 \\$\pm$0.3}&\thead{ 89.5 \\$\pm$0.3}&\thead{ 87.6 \\$\pm$0.2}&\thead{ 88.5 \\$\pm$0.2}&\thead{ 93.1 \\$\pm$0.1}&\thead{ 90.9 \\$\pm$0.3}&\thead{ 92.6 \\$\pm$0.1}&\thead{ 89.0 \\$\pm$0.2}&\thead{ 90.8 \\$\pm$0.2}\\
0.0196 &\thead{ 86.4 \\$\pm$0.2}&\thead{ 82.2 \\$\pm$0.3}&\thead{ 85.3 \\$\pm$0.3}&\thead{ 79.4 \\$\pm$0.6}&\thead{ 83.2 \\$\pm$0.3}&\thead{ 90.6 \\$\pm$0.4}&\thead{ 89.0 \\$\pm$0.3}&\thead{ 89.7 \\$\pm$0.3}&\thead{ 87.4 \\$\pm$0.4}&\thead{ 88.5 \\$\pm$0.4}&\thead{ 93.1 \\$\pm$0.1}&\thead{ 91.0 \\$\pm$0.2}&\thead{ 92.6 \\$\pm$0.2}&\thead{ 89.0 \\$\pm$0.2}&\thead{ 90.9 \\$\pm$0.2}\\
\underline{0.196} &\thead{ \textbf{86.6} \\$\pm$0.2}&\thead{ \textbf{84.0} \\$\pm$0.2}&\thead{ \textbf{86.5} \\$\pm$0.1}&\thead{ \textbf{81.7} \\$\pm$0.2}&\thead{ \textbf{84.0} \\$\pm$0.3}&\thead{ \textbf{91.4} \\$\pm$0.2}&\thead{ \textbf{90.7} \\$\pm$0.3}&\thead{ \textbf{91.1} \\$\pm$0.2}&\thead{ \textbf{89.8} \\$\pm$0.4}&\thead{ \textbf{90.1} \\$\pm$0.1}&\thead{ \textbf{93.6} \\$\pm$0.2}&\thead{ \textbf{92.1} \\$\pm$0.3}&\thead{ \textbf{93.3} \\$\pm$0.1}&\thead{ \textbf{91.1} \\$\pm$0.1}&\thead{ \textbf{92.0} \\$\pm$0.2}\\
\bottomrule
\end{tabular}
}
\caption{Accuracy\inpercent{}\tablefinisher{\gls{cLW}} for \resnet{}\bldStatement\undlinstmt}
\label{table:Accuracy in stdbeta}
\end{table*}
\begin{table*}
\resizebox{\linewidth}{!}{
\centering
\begin{tabular}{c|ccccc|ccccc|ccccc}
\toprule

0 &\thead{ 50.2 \\$\pm$0.2}&\thead{ 46.0 \\$\pm$0.2}&\thead{ 43.4 \\$\pm$0.3}&\thead{ 48.0 \\$\pm$0.5}&\thead{ 46.5 \\$\pm$0.3}&\thead{ 46.5 \\$\pm$0.5}&\thead{ 43.8 \\$\pm$0.7}&\thead{ 40.9 \\$\pm$0.4}&\thead{ 45.9 \\$\pm$0.6}&\thead{ 44.0 \\$\pm$0.6}&\thead{ 45.0 \\$\pm$0.2}&\thead{ 41.7 \\$\pm$0.3}&\thead{ 39.1 \\$\pm$0.1}&\thead{ 43.6 \\$\pm$0.6}&\thead{ 43.7 \\$\pm$0.4}\\
0.00196 &\thead{ 51.0 \\$\pm$0.1}&\thead{ 46.1 \\$\pm$0.2}&\thead{ 43.7 \\$\pm$0.2}&\thead{ 49.2 \\$\pm$0.6}&\thead{ 47.6 \\$\pm$0.4}&\thead{ 48.1 \\$\pm$0.4}&\thead{ 44.6 \\$\pm$0.4}&\thead{ 41.5 \\$\pm$0.4}&\thead{ 45.6 \\$\pm$0.5}&\thead{ 43.8 \\$\pm$0.3}&\thead{ 46.3 \\$\pm$0.0}&\thead{ 42.6 \\$\pm$0.3}&\thead{ 39.7 \\$\pm$0.2}&\thead{ 43.0 \\$\pm$0.7}&\thead{ 43.3 \\$\pm$0.5}\\
0.0196 &\thead{ 64.0 \\$\pm$0.6}&\thead{ 50.8 \\$\pm$0.6}&\thead{ 47.8 \\$\pm$0.4}&\thead{ 50.2 \\$\pm$1.0}&\thead{ 48.6 \\$\pm$0.9}&\thead{ 75.4 \\$\pm$0.5}&\thead{ 54.4 \\$\pm$0.7}&\thead{ 50.2 \\$\pm$0.6}&\thead{ 50.2 \\$\pm$1.9}&\thead{ 48.5 \\$\pm$1.0}&\thead{ 66.1 \\$\pm$0.7}&\thead{ 48.7 \\$\pm$0.6}&\thead{ 45.2 \\$\pm$0.4}&\thead{ 46.6 \\$\pm$2.0}&\thead{ 46.0 \\$\pm$1.3}\\
\underline{0.196} &\thead{ \textbf{98.9} \\$\pm$0.1}&\thead{ \textbf{69.9} \\$\pm$0.5}&\thead{ \textbf{71.9} \\$\pm$0.4}&\thead{ \textbf{65.2} \\$\pm$1.4}&\thead{ \textbf{72.6} \\$\pm$0.3}&\thead{ \textbf{98.8} \\$\pm$0.2}&\thead{ \textbf{85.7} \\$\pm$1.5}&\thead{ \textbf{86.6} \\$\pm$1.4}&\thead{ \textbf{69.3} \\$\pm$0.8}&\thead{ \textbf{73.9} \\$\pm$1.3}&\thead{ \textbf{99.2} \\$\pm$0.1}&\thead{ \textbf{72.4} \\$\pm$2.0}&\thead{ \textbf{74.6} \\$\pm$1.5}&\thead{ \textbf{63.7} \\$\pm$1.3}&\thead{ \textbf{74.8} \\$\pm$0.9}\\
\bottomrule
\end{tabular}
}
\caption{\loc{5}\inpercent{}\tablefinisher{\gls{cLW}} for \resnet{}\bldStatement\undlinstmt}
\label{table:loc5 in stdbeta}
\end{table*}

\begin{figure*}[t]
\begin{center}
  \includegraphics[width=.6\linewidth]{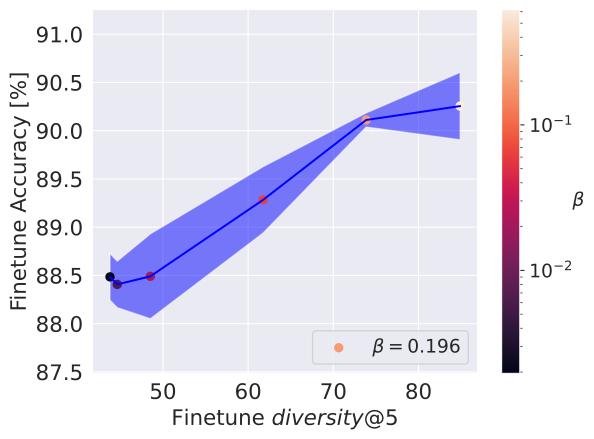}
\end{center}
  \caption{Relationship between finetuned \loc{5} and accuracy for varying \gls{cLW} for \resnet{}, portrayed via color, on \fgvcheader. Each dot represents an increase by a factor of $\sqrt{10}$ and the standard deviation is indicated by the shaded area. $\gls{cLW}=1.96$ \gls{customLoss} is not shown, as it dominates the training. }
   
\vspace{-0.55cm}
\label{fig:DivAcc}
\end{figure*}

\comparison{Scarlet Tanager}{0}
\comparison{White Crowned Sparrow}{3}
\comparison{Brown Creeper}{1}
\comparison{Cape Glossy Starling}{2}
\comparison{Ford Freestar Minivan}{4}
\comparison{Audi S5 Coupe 2012}{5}
\comparison{Ford Ranger SubCab2011}{6}
\comparison{Tornado}{7}
\comparison{Fokker 100}{8}

\end{document}